\documentclass[letterpaper]{article}
\usepackage{aaai2027arxiv}
\usepackage[hyphens]{url}
\usepackage{graphicx}
\usepackage{natbib}
\usepackage{caption}
\nocopyright

\usepackage{amsmath}
\usepackage{amssymb}
\usepackage{amsthm}
\usepackage{mathtools}
\usepackage{bm}
\usepackage{algorithm}
\usepackage{algorithmic}
\usepackage{booktabs}
\usepackage{array}
\usepackage{float}
\usepackage[many]{tcolorbox}
\usepackage{colortbl}
\usepackage{enumitem}
\allowdisplaybreaks
\usepackage{multirow}
\usepackage{tabularx}
\usepackage{multicol}
\usepackage{pifont}
\usepackage{setspace}

\usepackage{titlesec}
\usepackage{tocloft}
\usepackage{fancyhdr}

\usepackage[colorlinks,linkcolor=black,citecolor=black,urlcolor=black,bookmarksnumbered,unicode]{hyperref}
\hypersetup{pdftitle={Rectify Then Diffuse: Disentangling Concepts Before Denoising Trajectory Unfolds}}

\definecolor{rtdnavy}{HTML}{2B66B0}
\definecolor{rtdteal}{HTML}{147D78}
\definecolor{rtdpale}{HTML}{EEF5F8}
\definecolor{suppmid}{HTML}{E7F4F2}
\definecolor{rtdrule}{HTML}{CBD5E1}
\definecolor{rtdgray}{HTML}{CBD5E1}
\definecolor{rtdnavy}{HTML}{2B66B0}
\definecolor{rtdgray}{HTML}{F3F5F7}
\definecolor{rtdtealpale}{HTML}{E7F4F2}
\definecolor{citegreen}{HTML}{246B45}
\definecolor{urlblue}{HTML}{235B8E}
\definecolor{accentblue}{HTML}{2B66B0}
\definecolor{gainpos}{HTML}{147D78}
\definecolor{gainposdk}{HTML}{147D78}
\definecolor{gainneg}{HTML}{147D78}
\definecolor{gainnegdk}{HTML}{147D78}
\definecolor{rtdtealpale}{HTML}{E7F4F2}

\titleformat{\section}{\Large\bf\centering}{\thesection}{0.7em}{}
\titlespacing*{\section}{0pt}{2.0ex plus 0.5ex minus .2ex}{3pt plus 2pt minus 1pt}
\titleformat{\subsection}{\large\bf\raggedright}{\thesubsection}{0.7em}{}
\titlespacing*{\subsection}{0pt}{2.0ex plus 0.5ex minus .2ex}{3pt plus 2pt minus 1pt}
\titleformat{\subsubsection}[runin]{\normalsize\bf}{}{0pt}{}
\titlespacing*{\subsubsection}{0pt}{6pt plus 2pt minus 1pt}{1em}
\titleformat{\subparagraph}[runin]{\normalsize\bf}{}{0pt}{}
\titlespacing*{\subparagraph}{0pt}{6pt plus 2pt minus 1pt}{1em}
\titleformat{\paragraph}[runin]{\normalsize\bf}{}{0pt}{}
\titlespacing*{\paragraph}{0pt}{6pt plus 2pt minus 1pt}{1em}

\theoremstyle{plain}

\theoremstyle{definition}

\theoremstyle{remark}

\theoremstyle{plain}

\theoremstyle{definition}

\theoremstyle{remark}

\tcbset{
  graybox/.style={
    breakable,
    colback=black!10,
    colframe=white,
    width=\dimexpr\linewidth+10pt\relax,
    enlarge left by=-5pt,
    enlarge right by=-5pt,
    boxrule=0pt,
    left=0pt,right=0pt,top=0pt,bottom=0pt,
    sharp corners
  },
  thmbox/.style={
    breakable,
    enhanced jigsaw,
    colback=rtdpale,
    colframe=rtdnavy!72,
    boxrule=0pt,
    leftrule=2.2pt,
    arc=1.5pt,
    outer arc=1.5pt,
    boxsep=0pt,
    left=7pt,
    right=7pt,
    top=6pt,
    bottom=6pt,
    before skip=0.9\topsep,
    after skip=0.9\topsep,
  }
}
\newcommand{\csalwraptcb}[2]{
  \expandafter\let\csname csal@wrap@#1\expandafter\endcsname\csname #1\endcsname
  \expandafter\let\csname csal@wrap@end#1\expandafter\endcsname\csname end#1\endcsname
  \renewenvironment{#1}[1][]{
    \begin{tcolorbox}[#2]
    \if\relax\detokenize{##1}\relax
      \csname csal@wrap@#1\endcsname
    \else
      \csname csal@wrap@#1\endcsname[##1]
    \fi
  }{
    \csname csal@wrap@end#1\endcsname
    \end{tcolorbox}
  }
}
\csalwraptcb{assumption}{graybox, boxsep=5pt, before skip=\topsep, after skip=\topsep}
\csalwraptcb{theorem}{graybox, boxsep=3pt, before skip=4pt, after skip=4pt}
\csalwraptcb{lemma}{graybox, boxsep=5pt, before skip=\topsep, after skip=\topsep}

\DeclareMathOperator*{\argmax}{arg\,max}
\newcommand{\cmark}{\textcolor{rtdteal}{\ding{51}}}
\newcommand{\xmark}{\textcolor{black!55}{\ding{55}}}

\newcolumntype{Y}{>{\raggedright\arraybackslash}X}
\newcolumntype{C}{>{\centering\arraybackslash}X}
\newcolumntype{L}[1]{>{\raggedright\arraybackslash}p{#1}}

\newcommand{\clem}[1]{\hyperref[#1]{Lemma~\ref*{#1}}}
\newcommand{\cassum}[1]{\hyperref[#1]{Assumption~\ref*{#1}}}
\newcommand{\cdefn}[1]{\hyperref[#1]{Definition~\ref*{#1}}}

\title{Rectify Then Diffuse: Disentangling Concepts Before Denoising Trajectory Unfolds}
\author{
    Ning Zhu\textsuperscript{\rm 1},
    An Chen\textsuperscript{\rm 1},
    Mengfei Zhao,
    Juntao Xu\textsuperscript{\rm 1},
    Jingze Liang\textsuperscript{\rm 1},
    Boyuan Gu\textsuperscript{\rm 1},
    Liang-Jian Deng\textsuperscript{\rm 2}\thanks{Corresponding author.}
}
\affiliations{
    \textsuperscript{\rm 1}Glasgow College, University of Electronic Science and Technology of China\\
    \textsuperscript{\rm 2}School of Mathematical Sciences, University of Electronic Science and Technology of China
}

\makeatletter
\g@addto@macro\@maketitle{%
  {\centering
   \includegraphics[width=\textwidth]{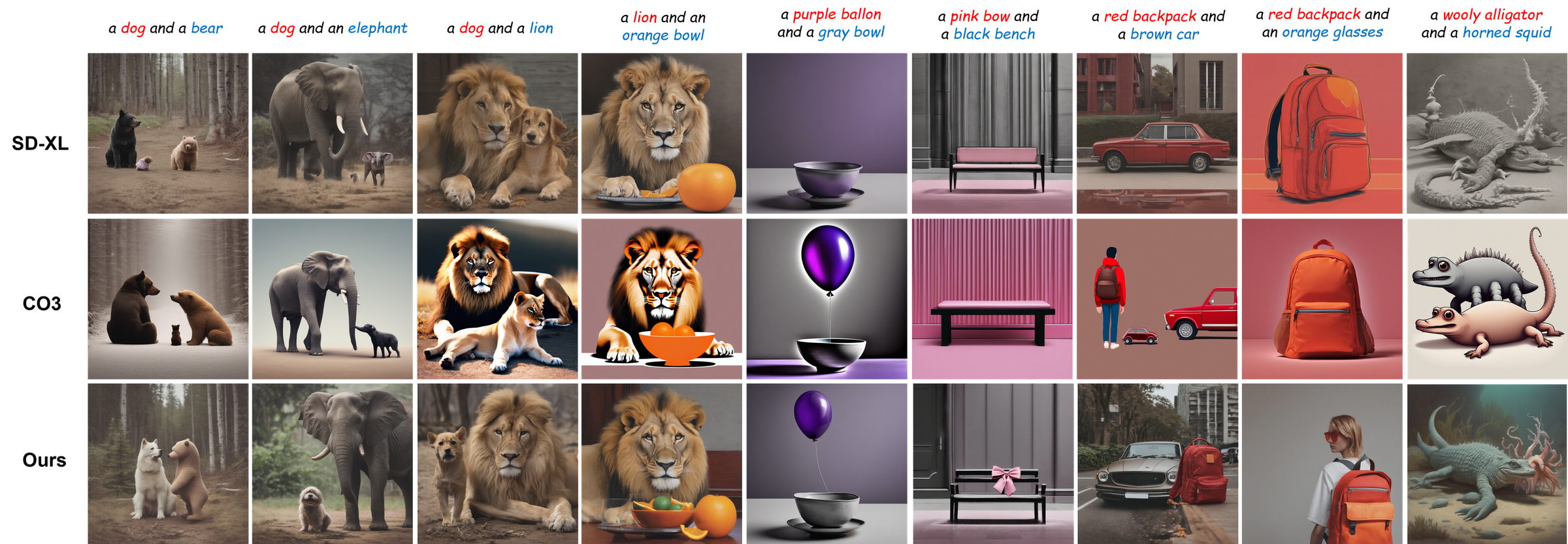}
   \captionof{figure}{Qualitative comparison on AE-Bench. From top to bottom: SDXL, CO3 and RTD.}
   \label{fig:qualitative}
   \par}
}
\makeatother

\begin{document}

\maketitle

\begin{abstract}
Text-to-image diffusion models can generate individual concepts well, but they often omit or merge concepts  incorrectly with multiple concepts. We trace these failures to an early coordination bottleneck: before denoising begins, prompt-conditioned attention may allocate different concepts to strongly overlapping spatial support, which can keep their attention coupled as denoising proceeds. This observation motivates treating compositional generation as a boundary-condition problem rather than repeatedly controlling the evolving trajectory. To this end, we propose \textbf{Rectify-then-Diffuse} (RTD), a training-free framework that rectifies the initial allocation once before standard denoising. Firstly, we propose \textbf{Soft-Overlap Disentanglement} (SOD), which converts normalized overlap between pilot concept maps into a differentiable and layout-agnostic separation objective. Secondly, we introduce \textbf{Isotropic Gradient Rectification} (IGR), which normalizes the SOD gradient and applies a bounded latent displacement with a consistent scale across prompts and initializations.  Extensive experiments show that RTD achieves state-of-the-art compositional fidelity and robust gains. On the AE-Bench object pair subset, RTD improves BLIP-VQA by 45.8\% and ImageReward by 19.6\% over CO3 while running 2.3$\times$ faster. Code will be released at \underline{\url{https://github.com/Z-yiwei/rectify-then-diffuse}}.
\end{abstract}

\section{Introduction}

Text-to-image diffusion models can produce convincing images of individual concepts~\cite{su2025higher,podell2024sdxl,esser2024sd3,ramesh2022dalle2,saharia2022imagen,nichol2022glide,chen2024pixartalpha}, yet their reliability drops when several concepts must coexist~\cite{hu2023tifa,ghosh2023geneval,cho2023visualprogramming,wu2024conceptmix,zhou2024migc,dat2025vsc}. A prompt as simple as ``a cat and a dog'' may yield only one animal, a hybrid of both animals, or two objects with incorrectly exchanged attributes~\cite{chefer2023attend,feng2023structure,liu2022composable,meral2024conform,lian2024llmgrounded}. These failures reveal a coordination bottleneck beyond the model's ability to represent each concept in isolation. Existing explanations commonly focus on imperfect text conditioning or incorrect semantic binding~\cite{zarei2024compositional}. We study a complementary spatial problem: although the model can represent every requested concept, it may initially allocate them to strongly overlapping regions of the latent canvas. When several concepts begin from the same spatial support, making later separation difficult.

Most training-free compositional methods intervene after denoising began. Attention-guidance methods strengthen neglected concepts or separate their attention maps over multiple sampling steps~\cite{chefer2023attend,agarwal2023astar,meral2024conform}, while corrective sampling methods repeatedly modify the evolving score or latent trajectory~\cite{park2025co3}. Other approaches introduce spatial supervision through predicted boxes, attention masks, or language-derived layouts~\cite{wang2024attentioncontrol,liu2026ucc,zhu2025csal}. These strategies improve fidelity but couple correction to evolving image structure. Once denoising starts, allocation and layout develop together, so recovering one concept may disturb structures that have already formed. InitNO moves the intervention to the initial latent, but it evaluates noise through several attention criteria and searches iteratively for a valid region~\cite{guo2024initno}. This leaves a more direct question unresolved: \textit{Is the conflict between concepts already observable before the denoising trajectory unfolds, and can it be corrected with a single targeted intervention?} To answer this question, we inspect the concept attention map produced by a high-noise pilot forward pass~\cite{hertz2023prompt2prompt,tang2023daam}. Although the initial latent does not yet contain recognizable image structure, its concept attention already exhibit non-random spatial organization. In unsuccessful generations, different concepts often concentrate on overlapping locations. Their attention can remain coupled during early denoising and later develop into omission or semantic fusion. In contrast, concepts with more distinct initial support tend to remain separable and appear together in the final image. Therefore, we identify early concept allocation as an actionable bottleneck whose state can be measured before scene formation. Because the overlap is available from start and is differentiable, it also provides a direct signal for changing boundary conditions of generation.

Based on this observation, we introduce \textbf{Rectify-then-Diffuse} (RTD), a training-free framework that corrects early concept allocation once and then returns generation to the original sampler. RTD first performs a diagnostic pilot pass to extract the provisional spatial support of each target concept. It then applies \textbf{Soft-Overlap Disentanglement} (SOD), which normalizes the concept maps and measures their pairwise soft overlap. SOD encourages different concepts to obtain distinct support without assigning target coordinates or object sizes. The resulting objective identifies a favorable direction for changing the initial latent, but its raw gradient magnitude can vary substantially. We therefore introduce \textbf{Isotropic Gradient Rectification} (IGR), which normalizes this gradient and constrains the displacement relative to the norm of the sampled latent. IGR turns the separation signal into a predictable one-shot correction. After this update, RTD performs standard denoising without further attention guidance, score correction, or solver-specific modification. This intervention is efficient since it only requires one additional forward pass. We evaluate RTD on three compositional benchmarks and RTD achieves the strongest overall performance, see Figure~\ref{fig:qualitative}. On the challenging object--object subset of AE-Bench, it reaches 0.7503 BLIP-VQA, improving over CO3 by 45.8\%. Moreover, RTD adds only 6.3\% overhead to standard SDXL inference and runs 2.3$\times$ faster than CO3 under the same setting.

\noindent\textbf{Our contributions are threefold:} (1) We identify early concept allocation as an actionable bottleneck, linking overlapping attention to coupled concept evolution and compositional failures. (2) We propose RTD, a training-free framework that combines layout-agnostic SOD with scale-stable IGR for initial latent rectification. (3) We demonstrate state-of-the-art compositional fidelity across three benchmarks, with consistent gains across backbones, solvers, and sampling budgets at only 6.3\% inference overhead.

\section{Related Work}

\begin{figure*}[t]
    \centering
    \includegraphics[width=\textwidth]{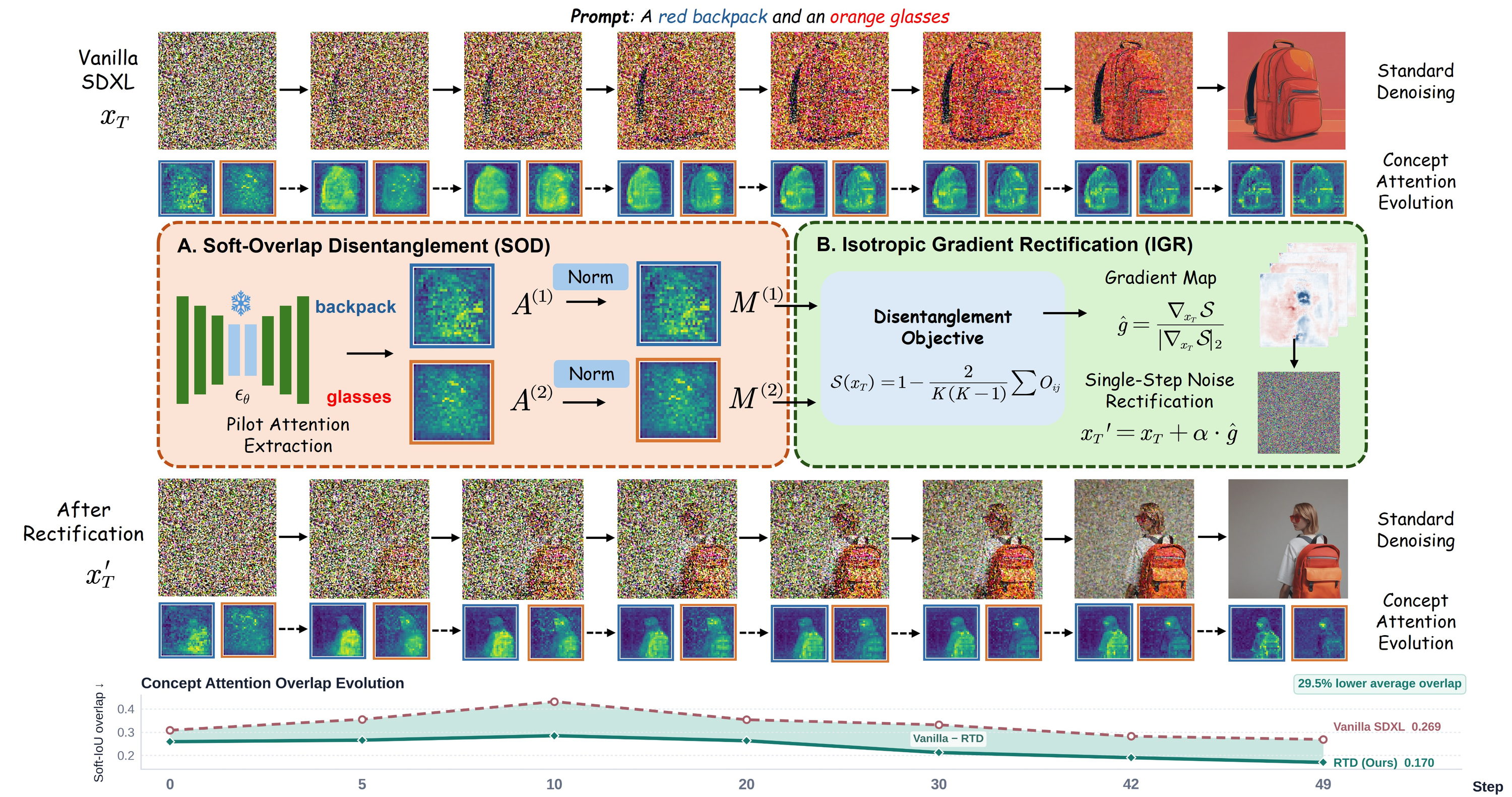}
    \caption{Overview of RTD. RTD extracts pilot concept maps, uses SOD to measure overlap, and applies IGR update. The rectified latent yields more distinct attention and preserves both concepts. The bottom plot tracks pairwise IoU.}
    \label{fig:method}
\end{figure*}

Compositional methods commonly modify attention, scores, or guidance during denoising to improve concept binding~\cite{chefer2023attend,rassin2023sygen,agarwal2023astar,liu2022composable,feng2023structure,kwon2024tweediemix,park2025co3,hertz2023prompt2prompt,chen2024layout,kim2023densediffusion,xie2023boxdiff,meral2024conform,lian2024llmgrounded,epstein2023selfguidance,hong2023sag,li2025mccd}. \cite{wang2024attentioncontrol} adds step-wise box prediction and attention masks, while UCC modulates early sampling using an LLM-derived spatial--semantic blueprint~\cite{liu2026ucc}. In multi-concept personalization, most recent works disentangle reference-image concepts through attention calibration, guided masks, or token adaptation with latent optimization~\cite{zhang2024disendiff,shentu2024attencraft,lim2025conceptsplit}. Before sampling, InitNO iteratively optimizes noise using multiple attention criteria and a distribution constraint~\cite{guo2024initno}.

\section{Methodology}
In this section, we provide a detailed description of our proposed RTD framework, see Figure~\ref{fig:method} for a brief overview. It includes three stages: pilot attention extraction, soft-overlap disentanglement and isotropic gradient rectification.

\subsection{Pilot Attention Extraction}
We first introduce pilot attention extraction to identify concept-allocation conflicts already encoded in the initial condition. Let $C$ be a prompt containing $K$ target concepts $\mathcal{C}=\{c_k\}_{k=1}^{K}$. Given a frozen denoising network $\epsilon_\theta$ and sampler $\Phi_\theta$, conventional generation samples $x_T\sim\mathcal{N}(0,\sigma_T^2I)$ and obtains $x_0=\Phi_\theta(x_T,C)$. RTD inserts a rectification operator before this sampling process:
\begin{equation}
    x_T'=\mathcal{R}_\theta(x_T,C),
    \qquad
    x_0'=\Phi_\theta(x_T',C).
    \label{eq:problem}
\end{equation}
The rectifier only changes the initial latent but not the parameters of the denoiser or sampler. The generated trajectory may consequently differ, but every transition along that trajectory still lies in the learned inference dynamics. The rectification signal is obtained from cross-attention, which connects prompt tokens to spatial features inside the denoising network. For a cross-attention head,
\begin{equation}
    A=\operatorname{softmax}\!\left(\frac{QK_C^\top}{\sqrt d}\right)
    \in\mathbb{R}^{(h w)\times L},
    \label{eq:crossattn}
\end{equation}
where $Q\in\mathbb{R}^{(h w)\times d}$ represents the spatial queries, $K_C\in\mathbb{R}^{L\times d}$ contains the keys of the $L$ prompt tokens, and the softmax is applied along the token dimension. For each concept $c_k$, we collect the attention maps of its associated tokens and sum them into a concept-level response. Maps from different heads and layers are spatially aligned and averaged, yielding $A^{(k)}\in\mathbb{R}^{h\times w}$. RTD computes $\{A^{(k)}\}_{k=1}^{K}$ using one conditional U-Net pass at a high-noise timestep $t_{\mathrm{pilot}}$. This pass is diagnostic: it uses the same $x_T$ and prompt as subsequent generation but does not advance the sampler or alter the latent. Although the image structure has not yet emerged, the resulting attention maps already exhibit prompt-dependent spatial organization. We interpret each $A^{(k)}$ as the provisional support assigned to concept $c_k$, rather than as a final segmentation mask. When several maps concentrate on the same locations, the corresponding concepts begin generation from conflicting spatial support. This observation makes their overlap a direct signal for rectifying the initial condition. Aggregating aligned maps across heads and layers reduces isolated attention responses and emphasizes allocation patterns that are consistently expressed by the frozen network.

\subsection{Soft-Overlap Disentanglement}
We then propose SOD to convert observed allocation conflicts into a differentiable objective. A suitable objective should compare the spatial support of different concepts while remaining insensitive to their absolute attention strengths. Direct overlap between the raw maps does not satisfy this requirement because attention ranges can vary across concepts and network layers. We therefore transform each $A^{(k)}$  into a soft occupancy map $M^{(k)}\in[0,1]^{h\times w}$ via max-min normalization.
Per-concept normalization preserves the relative spatial pattern within each map. For a pair of concepts $(c_i,c_j)$, SOD measures their shared support using a soft intersection-over-union:
\begin{equation}
    \mathrm{O}_{ij} = \frac{\langle M^{(i)}, M^{(j)} \rangle}
    {\|M^{(i)}\|_1+\|M^{(j)}\|_1-\langle M^{(i)}, M^{(j)} \rangle+\epsilon}.
    \label{eq:siou}
\end{equation}
$\mathrm{O}_{ij}$ increases when two concepts place strong responses at the same locations and approaches zero as their supports become more distinct. In contrast to overlap between thresholded masks, this measure retains graded attention information and permits gradients to propagate through the pilot pass. To handle multiple prompts, we average the overlap over all unordered pairs and define the separation objective:
\begin{equation}
    \mathcal{S}(x_T)=1-\frac{2}{K(K-1)}
    \sum_{1\leq i<j\leq K}\mathrm{O}_{ij}.
    \label{eq:sep}
\end{equation}
Maximizing $\mathcal{S}$ reduces pairwise competition for spatial support. The pairwise average treats all concepts symmetrically and keeps the scale of the objective comparable for different values of $K$. Since pilot attention maps are functions of $x_T$, the objective is differentiable and supplies a direction for improving its allocation. SOD imposes separation without constructing a layout. It does not assign coordinates, prescribe object sizes, or divide the canvas into predefined regions. Instead, it only discourages multiple concepts from relying on the same support. This distinction allows RTD to correct an early conflict while retaining the layout prior and generative capability of the original model.

\subsection{Isotropic Gradient Rectification}
We then propose IGR to turn the SOD objective into a predictable one-shot change of the initial latent. Let
$g=\nabla_{x_T}\mathcal{S}$ denote the gradient obtained by backpropagating through the pilot pass. Although $g$ identifies a locally favorable direction, its norm can vary considerably across prompts, concepts and sampled latents. A conventional update $x_T+\eta g$ therefore has an inconsistent effect: the same $\eta$ may barely alter one sample but excessively perturb another. IGR resolves this ambiguity by separating the direction of rectification from its magnitude. Specifically, we consider perturbations whose $\ell_2$ norm is bounded by $\rho\|x_T\|_2$, where the dimensionless parameter $\rho$ specifies a correction budget relative to the sampled latent norm. Under a first-order approximation of $\mathcal{S}$, the perturbation that maximizes its increase within this neighborhood follows the normalized gradient direction. This gives
\begin{equation}
    \hat{g}=\frac{g}{\max(\|g\|_2,\epsilon)},
    \qquad
    x_T'=x_T+\rho\|x_T\|_2\hat{g}.
    \label{eq:igr}
\end{equation}
The update magnitude is consequently controlled by the geometry of the initial latent. The ratio $\rho$ has the same interpretation across prompts: it determines how large the correction is relative to $\|x_T\|_2$. Because the constraint assigns an equal budget to every direction in latent space, we refer to this rectification as isotropic. The normalized update is invariant to positive rescaling of the objective gradient. IGR therefore preserves the direction identified by SOD while removing prompt- and seed-dependent variation in gradient magnitude. RTD performs this rectification once. It then discards the pilot objective and invokes the original sampler from $x_T'$, without applying further gradient guidance, attention constraints, or score modification during denoising.

\subsection{The RTD Algorithm and Design Properties}

We summarize the complete RTD procedure (see Algorithm~\ref{alg:rtd}) and its main design properties, RTD samples an initial latent, extracts the pilot concept maps, constructs the SOD objective from their pairwise overlap, and applies one IGR update before standard generation. The resulting framework has three properties. First, \textbf{the intervention is localized}: RTD modifies only the initial latent and does not introduce control operations into the denoising trajectory. Second, \textbf{the rectification cost is bounded}: it requires one additional forward pass and one backward pass, regardless of the number of subsequent inference steps. Third, \textbf{the method is sampler compatible}: because the solver-specific update rule remains unchanged, the same rectification procedure can precede different numerical solvers and sampling budgets.

\begin{algorithm}[t]
\caption{Rectify-then-Diffuse (RTD)}
\label{alg:rtd}
\begin{algorithmic}[1]
    \footnotesize
    \REQUIRE Prompt $C$, concepts $\mathcal{C}=\{c_k\}_{k=1}^{K}$, frozen network $\epsilon_\theta$, sampler $\Phi_\theta$, pilot timestep $t_{\mathrm{pilot}}$, relative correction ratio $\rho$
    \ENSURE Generated image $x_0'$
    \STATE $x_T\sim\mathcal{N}(0,\sigma_T^2I)$
    \STATE $\{A^{(k)}\}_{k=1}^{K}\leftarrow\operatorname{PilotAttention}(x_T,t_{\mathrm{pilot}},C,\epsilon_\theta)$
    \STATE $M^{(k)}\leftarrow\operatorname{Normalize}(A^{(k)}),\quad \forall k\in\{1,\dots,K\}$
    \FOR{$i\leftarrow1$ \TO $K-1$}
        \FOR{$j\leftarrow i+1$ \TO $K$}
            \STATE $\mathrm{O}_{ij}\leftarrow\dfrac{\langle M^{(i)},M^{(j)}\rangle}{\|M^{(i)}\|_1+\|M^{(j)}\|_1-\langle M^{(i)},M^{(j)}\rangle+\epsilon}$
        \ENDFOR
    \ENDFOR
    \STATE $\mathcal{S}\leftarrow1-\dfrac{2}{K(K-1)}\sum_{i<j}\mathrm{O}_{ij}$
    \STATE $g\leftarrow\nabla_{x_T}\mathcal{S}$
    \STATE $x_T'\leftarrow x_T+\rho\|x_T\|_2 g/\max(\|g\|_2,\epsilon)$
    \STATE $x_0'\leftarrow\Phi_\theta(x_T',C)$
    \RETURN $x_0'$
\end{algorithmic}
\end{algorithm}

\section{Experiments}
In this section, we introduce the experiments conducted on AE-Bench, T2I-CompBench, and RareBench to evaluate RTD for multi-concept compositional generation. We compare RTD with representative compositional generation methods and further examine its effectiveness and generality through ablation studies, sensitivity analysis, cross-backbone and cross-solver evaluation and runtime analysis.

\subsection{Experimental Setup}

\noindent\textbf{Implementation details.}
We use SDXL~\cite{podell2024sdxl} as the base model for all main experiments and generate images at a resolution of $1024\times1024$ using deterministic DDIM sampling~\cite{song2021ddim} with 50 inference steps. RTD performs one pilot forward pass and one backward pass at $t_{\mathrm{pilot}}=980$ with $\rho=0.02$, followed by standard sampling. Concepts are identified using Stanza~\cite{qi2020stanza} for RareBench, whitespace-based parsing for AE-Bench, and benchmark annotations for T2I-CompBench. For each method and prompt, we generate images using three fixed random seeds and report the mean performance. All experiments are conducted on a single NVIDIA H200 GPU.

\noindent\textbf{Datasets and metrics.}
We conduct our main evaluation on three benchmarks and follows the evaluation metrics used in CO3~\cite{park2025co3}. \textbf{AE-Bench}~\cite{chefer2023attend} contains 276 prompts from animal-animal (A-A), animal-object (A-O), and object-object (O-O) categories. We report BLIP-VQA~\cite{li2023blip2} and ImageReward~\cite{xu2023imagereward}. \textbf{T2I-CompBench}~\cite{huang2023t2icompbench} contains 1,000 prompts from its complex compositional split. \textbf{RareBench}~\cite{park2025co3} contains 120 long-tail prompts divided into Concat, Relations, and Complex categories. We report ImageReward (IR) and human evaluation (HE) for T2I-CompBench and RareBench. Human evaluation follows the protocol used by CO3~\cite{park2025co3} and adopted from R2F~\cite{kim2025r2f}. 11 participants assess prompt-image agreement using anonymized outputs presented in randomized order (see Appendix~\ref{app:experiments} for details).

\noindent\textbf{Baselines.}
We compare RTD with SD 1.5 and SDXL and with training-free baselines spanning attention guidance and semantic binding (Attend-and-Excite~\cite{chefer2023attend}, SynGen~\cite{rassin2023sygen}, Divide~\&~Bind~\cite{li2024dividebind}, Magnet~\cite{zhuang2024magnet}, and ToMe~\cite{hu2024tome}), compositional sampling (Composable Diffusion~\cite{liu2022composable} and TweedieMix~\cite{kwon2024tweediemix}), structured generation~\cite{kim2025r2f} (R2F), initial-noise optimization (InitNO)~\cite{guo2024initno}, and state-of-the-art corrective sampling (CO3)~\cite{park2025co3}.

\subsection{Main Results}
To quantify early concept competition, we further report the mean pairwise Step-5 Soft IoU computed from concept attention maps as SOD. We use Step~5 rather than Step~0 because methods that leave the initial noise unchanged are indistinguishable at initialization, while Step~5 remains early enough to allow method-specific allocation behavior to emerge.

\begin{figure}[!t]
    \centering
    \includegraphics[width=\columnwidth]{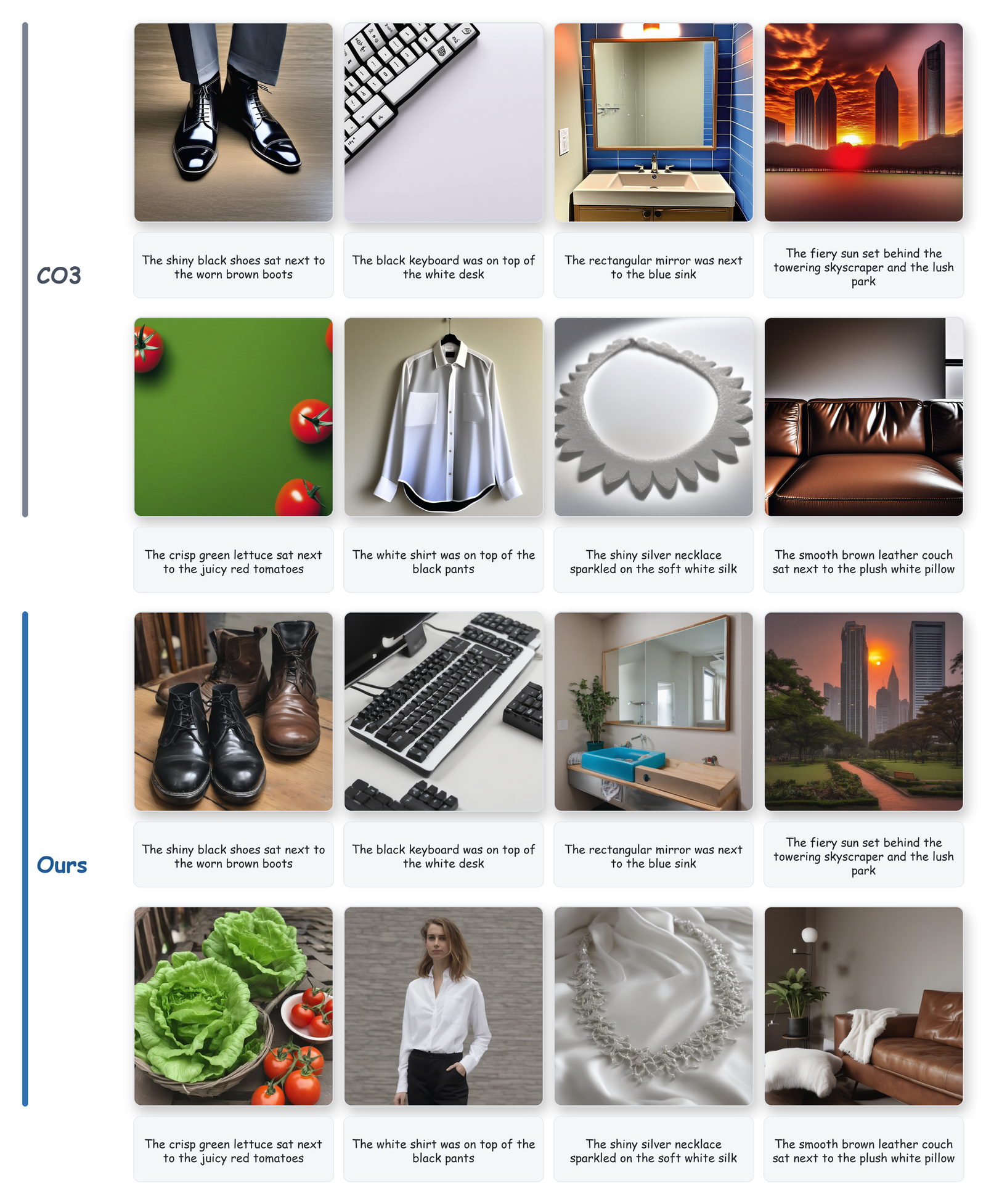}
    \caption{Qualitative comparison on T2I-CompBench. RTD more consistently renders both entities with the requested attributes and spatial relations than CO3.}
    \label{fig:exp_main_compbench}
\end{figure}

\begin{figure*}[htbp]
    \centering
    \includegraphics[width=\textwidth]{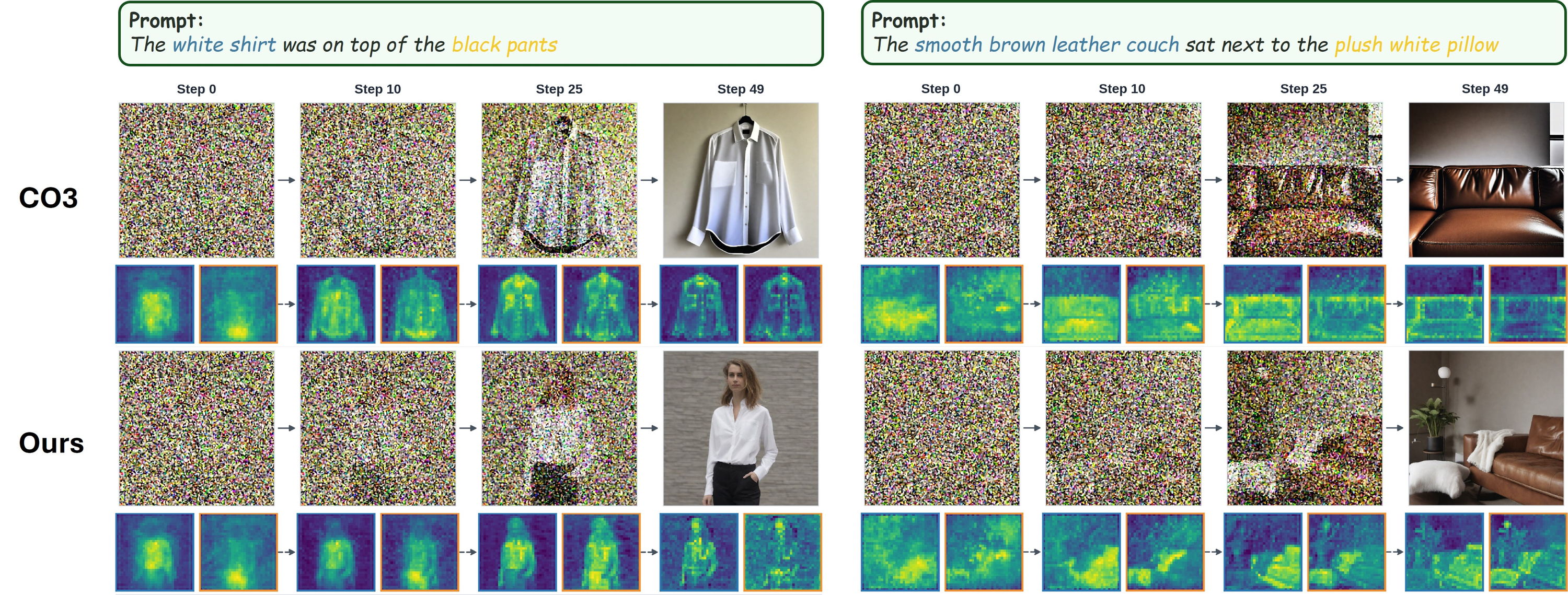}
    \caption{Concept attention evolution during denoising. Rows compare CO3 and RTD. Colored borders match the highlighted concepts. RTD separates their spatial support earlier and preserves both concepts in the final image.}
    \label{fig:visual_t2i}
\end{figure*}

\begin{figure}[t]
    \centering
    \includegraphics[width=\columnwidth]{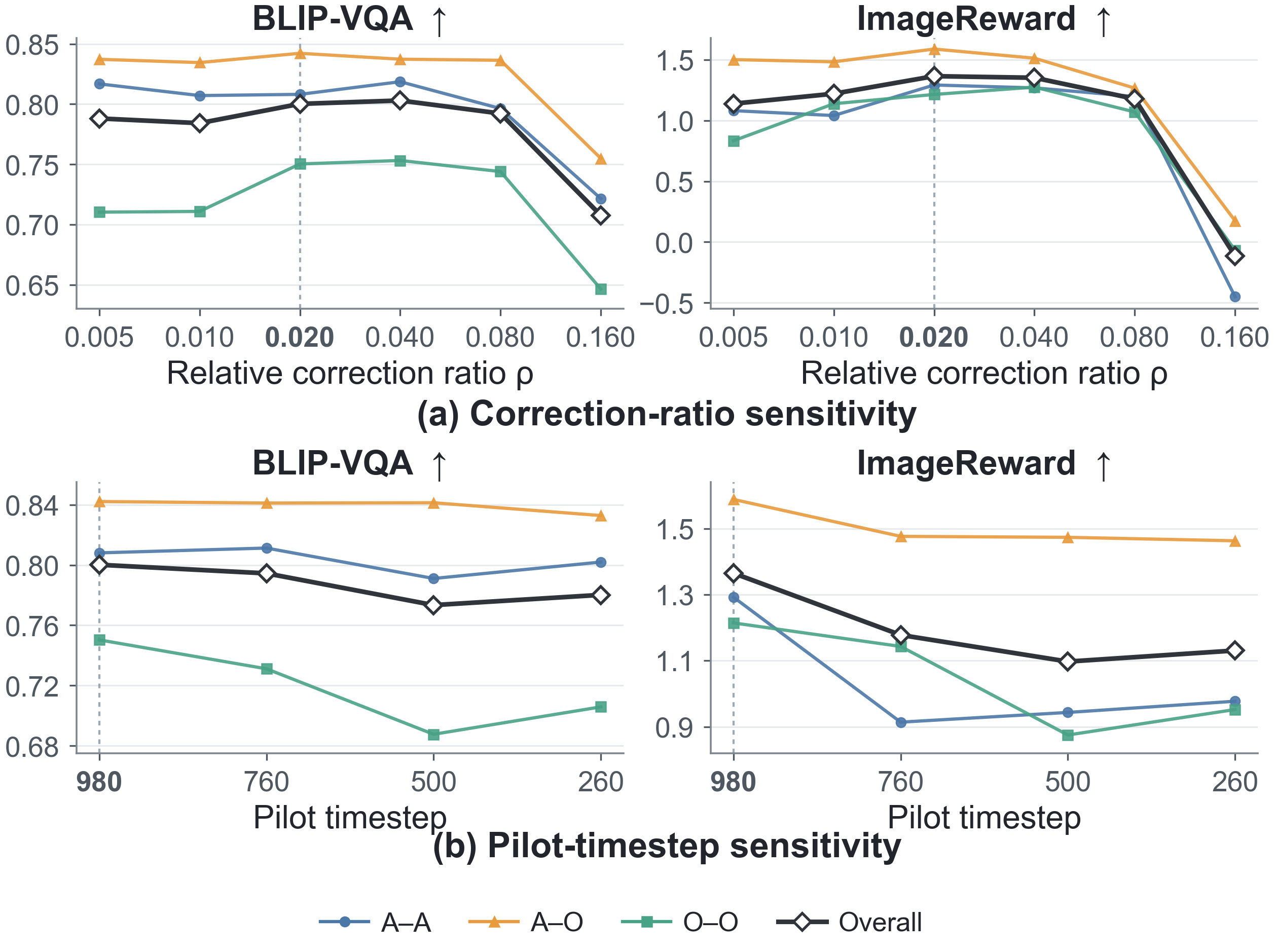}
    \caption{Sensitivity to (a) relative correction ratio $\rho$ and (b) pilot timestep $t_{\mathrm{pilot}}$ on AE-Bench.}
    \label{fig:sensitivity}
\end{figure}

\begin{table*}[t]
\centering
\caption{Quantitative results on AE-Bench. A-A, A-O, and O-O denote animal--animal, animal--object, and object--object.}
\label{tab:ae}
\footnotesize
\renewcommand{\arraystretch}{0.90}
\setlength{\tabcolsep}{3.5pt}

\begin{tabular}{l c c c c c c c c c}
\toprule
& & & \multicolumn{1}{c}{\textbf{Early Overlap}} & \multicolumn{3}{c}{\textbf{BLIP-VQA} ($\uparrow$)} & \multicolumn{3}{c}{\textbf{ImageReward} ($\uparrow$)} \\
\cmidrule(lr){4-4} \cmidrule(lr){5-7} \cmidrule(lr){8-10}
\textbf{Method} & \textbf{Training-Free} & \textbf{Model-Agnostic} & \textbf{S-IoU$_5$ $\downarrow$} & \textbf{A-A} & \textbf{A-O} & \textbf{O-O} & \textbf{A-A} & \textbf{A-O} & \textbf{O-O} \\
\midrule
SD 1.5            & \checkmark & - & 0.3476 & 0.3239 & 0.5958 & 0.2730 & $-0.2733$ & 0.4262 & $-0.5521$ \\
SDXL              & \checkmark & - & 0.2917 & 0.6950 & 0.8654 & 0.4926 & 0.7820 & 1.5574 & 0.6789 \\
Attend-and-Excite & \checkmark & $\times$   & 0.2734 & 0.6980 & 0.7865 & 0.5155 & 0.8244 & 1.2380 & 0.8741 \\
SynGen (SDXL)     & \checkmark & $\times$   & 0.2862 & 0.6816 & 0.8578 & 0.4652 & 0.6998 & 1.5622 & 0.6441 \\
Divide~\&~Bind    & \checkmark & $\times$   & 0.2598 & 0.7201 & 0.8399 & 0.5887 & 0.8499 & 1.2516 & 0.8134 \\
InitNO            & \checkmark & \checkmark   & 0.2521 & 0.7264 & 0.7998 & 0.5406 & 1.0082 & 1.3927 & 1.1383 \\
Magnet            & \checkmark & $\times$   & 0.2773 & 0.6782 & 0.8744 & 0.5398 & 0.6450 & 1.5895 & 0.7856 \\
ToMe              & \checkmark & $\times$   & 0.2649 & 0.6257 & 0.8808 & 0.6440 & 0.3895 & 1.5736 & 1.0118 \\
Comp.~Diff.       & \checkmark & \checkmark & 0.3318 & 0.2846 & 0.5656 & 0.4529 & $-1.1399$& $-0.2068$& $-0.0955$ \\
TweedieMix        & \checkmark & \checkmark & 0.2487 & 0.7390 & 0.8059 & 0.4683 & 1.0023 & 1.3127 & 0.7959 \\
CO3               & \checkmark & \checkmark & 0.2396 & 0.7441 & \textbf{0.8878} & 0.5146 & 1.2342 & \textbf{1.6744} & 1.0158 \\
\midrule
\cellcolor[gray]{0.90}\textbf{RTD (Ours)}
& \cellcolor[gray]{0.90}\checkmark
& \cellcolor[gray]{0.90}\checkmark
& \cellcolor[gray]{0.90}\textbf{0.2113}
& \cellcolor[gray]{0.90}\textbf{0.8081}
& \cellcolor[gray]{0.90}0.8422
& \cellcolor[gray]{0.90}\textbf{0.7503}
& \cellcolor[gray]{0.90}\textbf{1.2923}
& \cellcolor[gray]{0.90}1.5881
& \cellcolor[gray]{0.90}\textbf{1.2144} \\
\bottomrule
\end{tabular}
\end{table*}

\begin{table*}[!t]
\centering
\caption{Quantitative results on T2I-CompBench and RareBench.}
\label{tab:t2i_rare}
\footnotesize
\renewcommand{\arraystretch}{0.92}
\setlength{\tabcolsep}{3.2pt}
\begin{tabular}{l c c c c c c c c c c}
\toprule
& \multicolumn{3}{c}{\textbf{T2I-CompBench}} & \multicolumn{7}{c}{\textbf{RareBench}} \\
\cmidrule(lr){2-4} \cmidrule(lr){5-11}
& \multicolumn{2}{c}{\textbf{Complex}} & \multicolumn{1}{c}{\textbf{Early Overlap}} & \multicolumn{2}{c}{\textbf{Concat}} & \multicolumn{2}{c}{\textbf{Relations}} & \multicolumn{2}{c}{\textbf{Complex}} & \multicolumn{1}{c}{\textbf{Early Overlap}} \\
\cmidrule(lr){2-3} \cmidrule(lr){4-4} \cmidrule(lr){5-6} \cmidrule(lr){7-8} \cmidrule(lr){9-10} \cmidrule(lr){11-11}
\textbf{Method} & \textbf{IR $\uparrow$} & \textbf{HE $\uparrow$} & \textbf{S-IoU$_5$ $\downarrow$} & \textbf{IR $\uparrow$} & \textbf{HE $\uparrow$} & \textbf{IR $\uparrow$} & \textbf{HE $\uparrow$} & \textbf{IR $\uparrow$} & \textbf{HE $\uparrow$} & \textbf{S-IoU$_5$ $\downarrow$} \\
\midrule
SDXL          & 0.3083 & 0.4678 & 0.2831 & 0.1018    & 0.4590 & 0.3607    & 0.4133 & 1.1907 & 0.6174 & 0.2796 \\
SynGen (SDXL) & 0.2899 & 0.4127 & 0.2748 & 0.0576    & 0.3755     & 0.3177    & 0.4378     & 1.1452 & 0.5804 & 0.2709 \\
ToMe          & 0.2891 & 0.4361 & 0.2617 & $-0.2041$ & 0.3974 & $-0.0632$ & 0.4607 & 1.0200 & 0.4653 & 0.2852 \\
R2F           & 0.3561 & 0.3749 & 0.2536 & $-0.0162$ & 0.3862 & 0.3620    & 0.4796 & 1.1542 & 0.5536 & 0.2564 \\
CO3           & 0.4406 & 0.6278 & 0.2369 & 0.1610    & 0.5246 & 0.2588    & 0.4889 & 1.2211 & 0.5977 & 0.2317 \\
\midrule
\cellcolor[gray]{0.90}\textbf{RTD (Ours)}
& \cellcolor[gray]{0.90}\textbf{0.4661}
& \cellcolor[gray]{0.90}\textbf{0.6615}
& \cellcolor[gray]{0.90}\textbf{0.2146}
& \cellcolor[gray]{0.90}\textbf{0.4464}
& \cellcolor[gray]{0.90}\textbf{0.5869}
& \cellcolor[gray]{0.90}\textbf{0.4760}
& \cellcolor[gray]{0.90}\textbf{0.5117}
& \cellcolor[gray]{0.90}\textbf{1.2476}
& \cellcolor[gray]{0.90}\textbf{0.6204}
& \cellcolor[gray]{0.90}\textbf{0.2078} \\
\bottomrule
\end{tabular}
\end{table*}

\noindent\textbf{AE-Bench.}
Table~\ref{tab:ae} shows that RTD achieves state-of-the-art performance in four of the six category--metric settings and obtains the best BLIP-VQA and ImageReward on both A-A and O-O. On the particularly challenging O-O subset, RTD reaches 0.7503 BLIP-VQA and 1.2144 ImageReward, exceeding the previous best results by 16.5\% and 6.7\%. Relative to CO3, the gains reach 45.8\% in BLIP-VQA and 19.6\% in ImageReward. RTD also improves over CO3 by 8.6\% and 4.7\% on A-A, establishing its ability to jointly render concepts with closely related semantics.

\noindent\textbf{T2I-CompBench and RareBench.}
Table~\ref{tab:t2i_rare} establishes state-of-the-art performance on complex and long-tail compositions. On T2I-CompBench, RTD achieves the best ImageReward and human-evaluation scores of 0.4661 and 0.6615, improving over CO3 by 5.8\% and 5.4\%. On RareBench, RTD ranks first in every category and metric, including the Relations subset. It improves ImageReward over CO3 by 0.2854 on Concat and 0.2172 on Relations, while also achieving the strongest human-evaluation score in all three categories. The agreement between automatic and human evaluation demonstrates that RTD generalizes across complex structures, semantic relations, and long-tail concepts. RTD's consistently lowest S-IoU$_5$ indicates early feature-level concept separation, which is positively associated with its stronger multi-concept generation scores.

\noindent\textbf{Qualitative results.}
Figures~\ref{fig:qualitative} and~\ref{fig:exp_main_compbench} present qualitative comparisons on AE-Bench and T2I-CompBench. SDXL and CO3 frequently omit one requested concept, merge multiple entities, or bind attributes incorrectly. In contrast, RTD more consistently renders all target concepts while preserving their respective colors, categories, and spatial relations. For prompts involving a red backpack with orange glasses or a white shirt with black pants, RTD produces clearly separated entities with the requested attributes. Figure~\ref{fig:visual_t2i} compares the concept attention evolution of CO3 and RTD during denoising. Under CO3, paired attention maps remain concentrated in the same regions and are progressively dominated by one concept. RTD establishes distinct spatial support early in denoising and preserves this separation throughout the remaining sampling steps. The corresponding outputs retain both target concepts, providing direct visual evidence that the separation promoted by SOD leads to stronger compositional fidelity.
\subsection{Ablation Study}
Table~\ref{tab:abl_ae} evaluates the contributions of SOD and IGR, together with the number of rectification updates. Removing rectification recovers vanilla SDXL, while removing IGR applies the SOD gradient without normalization. The complete RTD method combines both components with a single update.

\begin{table*}[!t]
\centering
\caption{Ablation of RTD components and rectification updates on AE-Bench. $N_{\mathrm{rect}}$ denotes the number of updates.}
\label{tab:abl_ae}
\footnotesize
\renewcommand{\arraystretch}{0.88}
\begin{tabular}{l c c c c c c c c c}
\toprule
& \multicolumn{3}{c}{\textbf{Components}} & \multicolumn{3}{c}{\textbf{BLIP-VQA} ($\uparrow$)} & \multicolumn{3}{c}{\textbf{ImageReward} ($\uparrow$)} \\
\cmidrule(lr){2-4} \cmidrule(lr){5-7} \cmidrule(lr){8-10}
\textbf{Configuration} & \textbf{SOD} & \textbf{IGR} & $\boldsymbol{N_{\mathrm{rect}}}$ & \textbf{A-A} & \textbf{A-O} & \textbf{O-O} & \textbf{A-A} & \textbf{A-O} & \textbf{O-O} \\
\midrule
No rectification    & $\times$   & $\times$   & 0 & 0.6950 & \textbf{0.8654} & 0.4926 & 0.7820 & 1.5574 & 0.6789 \\
Unnormalized update & \checkmark & $\times$   & 1 & 0.7993 & 0.8288 & 0.7108 & 0.8703 & 1.4502 & 0.7946 \\
Two updates         & \checkmark & \checkmark & 2 & 0.8056 & 0.8447 & 0.7481 & 1.0913 & 1.5870 & 0.8767 \\
Four updates        & \checkmark & \checkmark & 4 & \textbf{0.8321} & 0.8299 & 0.7319 & 0.9829 & 1.4146 & 0.8268 \\
\midrule
\cellcolor[gray]{0.90}\textbf{RTD (Ours)}
& \cellcolor[gray]{0.90}\checkmark
& \cellcolor[gray]{0.90}\checkmark
& \cellcolor[gray]{0.90}1
& \cellcolor[gray]{0.90}0.8081
& \cellcolor[gray]{0.90}0.8422
& \cellcolor[gray]{0.90}\textbf{0.7503}
& \cellcolor[gray]{0.90}\textbf{1.2923}
& \cellcolor[gray]{0.90}\textbf{1.5881}
& \cellcolor[gray]{0.90}\textbf{1.2144} \\
\bottomrule
\end{tabular}
\end{table*}

\noindent\textbf{Component effectiveness.}
Compared with no rectification, RTD improves BLIP-VQA by 0.1131 on A-A and 0.2577 on O-O, while increasing ImageReward by 0.5103 and 0.5355. IGR further improves all six metrics over the unnormalized SOD update. The ImageReward gains reach 0.4220 on A-A, 0.1379 on A-O, and 0.4198 on O-O. These results confirm that the separation objective provides an effective correction signal and that gradient normalization is essential for translating it into reliable latent updates.

\noindent\textbf{Rectification updates.}
A single update achieves the best ImageReward in all three categories and the best O-O BLIP-VQA. Increasing the update count provides no consistent improvement. RTD therefore adopts $N_{\mathrm{rect}}{=}1$, achieving strong compositional fidelity with the minimum rectification cost.

\subsection{Sensitivity Study}

\noindent\textbf{Relative correction ratio.}
Figure~\ref{fig:sensitivity}(a) shows consistently strong performance across $\rho\in[0.02,0.08]$, with $\rho=0.02$ and $0.04$ providing the strongest overall results. We use $\rho=0.02$ because it achieves the best aggregate balance across AE-Bench categories and metrics. The broad high-performance range confirms that IGR provides a stable relative correction scale.

\noindent\textbf{Pilot timestep.}
Figure~\ref{fig:sensitivity}(b) shows that the highest-noise setting, $t_{\mathrm{pilot}}=980$, achieves the best ImageReward and the best O-O BLIP-VQA. This confirms that concept allocation is already actionable before substantial scene formation and supports rectification at the start of generation.

\begin{table*}[!t]
\centering
\caption[Cross-solver and cross-step performance on AE-Bench.]{Cross-solver and cross-step performance on AE-Bench.\footnotemark}

\label{tab:solver}
\footnotesize
\resizebox{\textwidth}{!}{
\renewcommand{\arraystretch}{0.88}
\begin{tabular}{l c c c c c c c c c c c c c}
\toprule
& & \multicolumn{6}{c}{\textbf{DDIM}} & \multicolumn{6}{c}{\textbf{DPM++~2M}} \\
\cmidrule(lr){3-8} \cmidrule(lr){9-14}
& & \multicolumn{3}{c}{\textbf{ImageReward} ($\uparrow$)} & \multicolumn{3}{c}{\textbf{BLIP-VQA} ($\uparrow$)} & \multicolumn{3}{c}{\textbf{ImageReward} ($\uparrow$)} & \multicolumn{3}{c}{\textbf{BLIP-VQA} ($\uparrow$)} \\
\cmidrule(lr){3-5} \cmidrule(lr){6-8} \cmidrule(lr){9-11} \cmidrule(lr){12-14}
\textbf{Method} & \textbf{Steps} & \textbf{A-A} & \textbf{A-O} & \textbf{O-O} & \textbf{A-A} & \textbf{A-O} & \textbf{O-O} & \textbf{A-A} & \textbf{A-O} & \textbf{O-O} & \textbf{A-A} & \textbf{A-O} & \textbf{O-O} \\
\midrule
SDXL & 10 & 0.4906 & 1.1503 & 0.1131 & 0.6428 & 0.8378 & 0.4752 & 0.6834 & \textbf{1.5074} & 0.6681 & 0.6768 & 0.8537 & 0.4974 \\
SDXL & 20 & 0.6965 & 1.4567 & 0.5516 & 0.6832 & \textbf{0.8620} & 0.4947 & 0.7089 & \textbf{1.5652} & 0.6757 & 0.6974 & 0.8537 & 0.4984 \\
SDXL & 50 & 0.7820 & 1.5574 & 0.6789 & 0.6950 & 0.8654 & 0.4926 & 0.7820 & 1.5574 & 0.6789 & 0.6950 & 0.8658 & 0.4925 \\
\midrule
CO3 & 10 & 0.5872 & \textbf{1.5900} & \textbf{0.7117} & 0.6294 & \textbf{0.8589} & 0.4453 & \textbf{1.0371} & 1.4520 & 0.9206 & 0.7331 & \textbf{0.8765} & 0.4443 \\
CO3 & 20 & 0.8532 & \textbf{1.6440} & 0.7481 & 0.6698 & 0.8611 & 0.4315 & \textbf{1.1424} & 1.5249 & \textbf{0.9537} & 0.7382 & \textbf{0.8842} & 0.4944 \\
CO3 & 50 & 1.2341 & \textbf{1.6743} & 1.0158 & 0.7441 & \textbf{0.8878} & 0.5146 & 1.2341 & \textbf{1.6743} & 1.0158 & 0.7441 & \textbf{0.8878} & 0.5146 \\
\midrule
\cellcolor[gray]{0.90}\textbf{RTD}
& \cellcolor[gray]{0.90}10
& \cellcolor[gray]{0.90}\textbf{0.7436}
& \cellcolor[gray]{0.90}1.1762
& \cellcolor[gray]{0.90}0.7095
& \cellcolor[gray]{0.90}\textbf{0.7895}
& \cellcolor[gray]{0.90}0.8334
& \cellcolor[gray]{0.90}\textbf{0.7084}
& \cellcolor[gray]{0.90}0.9722
& \cellcolor[gray]{0.90}1.3992
& \cellcolor[gray]{0.90}\textbf{0.9295}
& \cellcolor[gray]{0.90}\textbf{0.7895}
& \cellcolor[gray]{0.90}0.8334
& \cellcolor[gray]{0.90}\textbf{0.7084} \\
\cellcolor[gray]{0.90}\textbf{RTD}
& \cellcolor[gray]{0.90}20
& \cellcolor[gray]{0.90}\textbf{1.0524}
& \cellcolor[gray]{0.90}1.5644
& \cellcolor[gray]{0.90}\textbf{0.9192}
& \cellcolor[gray]{0.90}\textbf{0.8015}
& \cellcolor[gray]{0.90}0.8546
& \cellcolor[gray]{0.90}\textbf{0.6907}
& \cellcolor[gray]{0.90}1.1196
& \cellcolor[gray]{0.90}1.5629
& \cellcolor[gray]{0.90}0.9432
& \cellcolor[gray]{0.90}\textbf{0.8015}
& \cellcolor[gray]{0.90}0.8546
& \cellcolor[gray]{0.90}\textbf{0.6907} \\
\cellcolor[gray]{0.90}\textbf{RTD}
& \cellcolor[gray]{0.90}50
& \cellcolor[gray]{0.90}\textbf{1.2923}
& \cellcolor[gray]{0.90}1.5881
& \cellcolor[gray]{0.90}\textbf{1.2144}
& \cellcolor[gray]{0.90}\textbf{0.8081}
& \cellcolor[gray]{0.90}0.8422
& \cellcolor[gray]{0.90}\textbf{0.7503}
& \cellcolor[gray]{0.90}\textbf{1.2414}
& \cellcolor[gray]{0.90}1.6508
& \cellcolor[gray]{0.90}\textbf{1.0798}
& \cellcolor[gray]{0.90}\textbf{0.8131}
& \cellcolor[gray]{0.90}0.8742
& \cellcolor[gray]{0.90}\textbf{0.7476} \\
\bottomrule
\end{tabular}
}
\end{table*}

\footnotetext{All CO3 and SDXL results in this table are taken directly from the original paper~\cite{park2025co3}.}

\subsection{Cross-Backbone/Solver Generalization}
\label{sec:generalization}

\noindent\textbf{Cross-backbone.}
We apply RTD to SDXL and SD~2.1 with the same rectification rule and each backbone's native resolution. As shown in Table~\ref{tab:backbone_ae}, RTD achieves the best overall BLIP-VQA on both backbones. It raises the SDXL score from 0.7155 for CO3 to 0.8002 and obtains the strongest A-A and O-O results. On SD~2.1, RTD reaches 0.7110 overall and outperforms CO3 in every category. This consistent transfer establishes that RTD generalizes across diffusion backbones without backbone-specific tuning.

\noindent\textbf{Cross-solver and cross-step.}
We further evaluate DDIM and DPM++~2M~\cite{lu2022dpm} with 10, 20, and 50 inference steps while keeping RTD fixed. Across both solvers and every sampling budget, RTD achieves the strongest A-A and O-O BLIP-VQA. With only 10 steps, it reaches 0.7895 and 0.7084, already surpassing the 50-step CO3 scores of 0.7441 and 0.5146. At 50 steps, RTD also achieves the best A-A and O-O ImageReward under both solvers. These results establish solver-independent and budget-robust compositional gains without retuning.

\subsection{Time Efficiency}

Table~\ref{tab:runtime} compares per-image inference time on a single H200 GPU using SDXL at $1024\times1024$ resolution and 50-step DDIM sampling. RTD adds only 0.31\,s to the 4.89\,s vanilla pipeline, corresponding to a 6.3\% overhead. Its total runtime is 5.20\,s, which is 2.3$\times$ faster than CO3 under the same setting. This efficiency follows from performing rectification once at initialization and then reusing the original denoising pipeline. RTD therefore delivers substantial compositional gains while retaining near-baseline inference latency.

\begin{table}[!t]
\centering
\caption{AE-Bench BLIP-VQA across backbones ($\uparrow$).}
\label{tab:backbone_ae}
\footnotesize
\renewcommand{\arraystretch}{0.92}
\begin{tabular}{l l c c c c}
\toprule
\textbf{Model} & \textbf{Method} & \textbf{A-A} & \textbf{A-O} & \textbf{O-O} & \textbf{Overall} \\
\midrule
\multirow{3}{*}{SDXL}  & Vanilla & 0.6951 & 0.8654 & 0.4926 & 0.6844 \\
                        & CO3     & 0.7441 & \textbf{0.8878} & 0.5146 & 0.7155 \\
\cellcolor[gray]{0.90}
                        & \cellcolor[gray]{0.90}\textbf{RTD}
                        & \cellcolor[gray]{0.90}\textbf{0.8081}
                        & \cellcolor[gray]{0.90}0.8422
                        & \cellcolor[gray]{0.90}\textbf{0.7503}
                        & \cellcolor[gray]{0.90}\textbf{0.8002} \\
\midrule
\multirow{3}{*}{SD 2.1} & Vanilla & 0.6943 & \textbf{0.8003} & \textbf{0.6237} & 0.7061 \\
                         & CO3     & 0.7082 & 0.7554 & 0.5316 & 0.6650 \\
\cellcolor[gray]{0.90}
                         & \cellcolor[gray]{0.90}\textbf{RTD}
                         & \cellcolor[gray]{0.90}\textbf{0.7194}
                         & \cellcolor[gray]{0.90}0.7954
                         & \cellcolor[gray]{0.90}0.6181
                         & \cellcolor[gray]{0.90}\textbf{0.7110} \\
\bottomrule
\end{tabular}
\end{table}

\begin{center}
\begin{minipage}{\columnwidth}
\centering
\captionof{table}{Per-image runtime on one H200 GPU. Pilot/Corr. denotes pilot or corrective processing time.}
\label{tab:runtime}
\footnotesize
\renewcommand{\arraystretch}{0.92}
\begin{tabular}{l c c c}
\toprule
\textbf{Method} & \textbf{Pilot/Corr.} & \textbf{DDIM} & \textbf{Total} \\
\midrule
SDXL     & --                 & 4.89\,s & 4.89\,s \\
CO3      & (in DDIM)          & 12.02\,s & 12.02\,s \\
\cellcolor[gray]{0.90}\textbf{RTD}
& \cellcolor[gray]{0.90}0.31\,s
& \cellcolor[gray]{0.90}4.89\,s
& \cellcolor[gray]{0.90}\textbf{5.20\,s} \\
\bottomrule
\end{tabular}
\end{minipage}
\end{center}

\section{Limitations}
RTD is limited around a general design principle: different concepts should first obtain identifiable spatial support before their interaction is rendered. This layout-agnostic formulation already achieves state-of-the-art performance on the RareBench Relations subset. A promising extension is to condition SOD on relation semantics, allowing the rectification objective to explicitly model interaction-specific geometry such as holding, wearing, and occlusion. Such relation-aware rectification would extend the same initial-allocation framework to increasingly structured compositions.

\section{Conclusion}
We revisit multi-concept generation through semantic allocation before generation unfolds. Strong overlap among pilot concept attention maps can couple concepts throughout denoising and impede their separation. RTD therefore corrects the initial latent once before retaining the original generation pipeline: SOD converts soft attention overlap into a differentiable competition signal, and IGR produces a stable, controlled rectification. Across three compositional benchmarks, RTD achieves state-of-the-art performance. Consistent gains across backbones, solvers, and sampling budgets establish early semantic allocation as a general intervention point beyond any specific generative process. We hope this perspective inspires future methods that improve compositional generation by shaping initial conditions.

\onecolumn
\raggedbottom
\appendix
\counterwithin{figure}{section}
\counterwithin{table}{section}
\counterwithin{equation}{section}
\hypersetup{citecolor=citegreen,linkcolor=rtdnavy,urlcolor=urlblue}
\titleformat{\section}
  {\normalfont\Large\bfseries\color{rtdnavy}}
  {\thesection}{0.65em}{}
  [\vspace{1.5pt}\color{rtdrule}\titlerule]
\titleformat{\subsection}
  {\normalfont\large\bfseries\color{black!88}}
  {\thesubsection}{0.55em}{}
\titleformat{\subsubsection}
  {\normalfont\normalsize\bfseries\color{black!82}}
  {\thesubsubsection}{0.5em}{}
\titleformat{\paragraph}[runin]
  {\normalfont\normalsize\bfseries\color{rtdteal}}
  {}{0pt}{}[.]
\titlespacing*{\section}{0pt}{15pt plus 2pt minus 2pt}{8pt}
\titlespacing*{\subsection}{0pt}{10pt plus 2pt minus 1pt}{4pt}
\titlespacing*{\subsubsection}{0pt}{8pt plus 1pt minus 1pt}{3pt}
\titlespacing*{\paragraph}{0pt}{5pt}{0.45em}
\captionsetup{
  font=small,
  labelfont={bf,color=rtdnavy},
  textfont={color=black!88},
  justification=raggedright,
  singlelinecheck=false,
  skip=6pt
}
\captionsetup[table]{position=above,skip=5pt}
\captionsetup[figure]{position=below,skip=6pt}
\renewcommand{\topfraction}{0.95}
\renewcommand{\bottomfraction}{0.90}
\renewcommand{\textfraction}{0.05}
\renewcommand{\floatpagefraction}{0.75}
\setcounter{topnumber}{4}
\setcounter{bottomnumber}{3}
\setcounter{totalnumber}{6}
\setlength{\tabcolsep}{5pt}
\renewcommand{\arraystretch}{1.10}
\setstretch{1.025}
\setlength{\parindent}{1em}
\setlength{\parskip}{0.12em}
\setlength{\emergencystretch}{2em}
\setlength{\headheight}{13pt}
\setlength{\headsep}{16pt}
\addtolength{\textheight}{-29pt}
\fancyhf{}
\fancyhead[L]{\small\textcolor{black!68}{RTD: Supplementary Material}}
\fancyhead[R]{\small\textcolor{black!68}{\nouppercase{\rightmark}}}
\fancyfoot[R]{\small\textcolor{black!68}{\thepage}}
\renewcommand{\headrulewidth}{0.35pt}
\renewcommand{\footrulewidth}{0pt}
\renewcommand{\sectionmark}[1]{\markboth{#1}{#1}}
\fancypagestyle{suppfirst}{%
  \fancyhf{}
  \fancyfoot[R]{\small\textcolor{black!68}{\thepage}}
  \renewcommand{\headrulewidth}{0pt}
  \renewcommand{\footrulewidth}{0pt}
}
\pagestyle{fancy}
\renewcommand{\contentsname}{Contents}
\renewcommand{\cfttoctitlefont}{\large\bfseries\color{rtdnavy}}
\renewcommand{\cftsecfont}{\bfseries\color{black!85}}
\renewcommand{\cftsecpagefont}{\bfseries\color{black!70}}
\renewcommand{\cftsubsecfont}{\color{black!76}}
\renewcommand{\cftsubsecpagefont}{\color{black!62}}
\setlength{\cftbeforesecskip}{3pt}

\phantomsection
\addcontentsline{toc}{section}{Supplementary Material}
\thispagestyle{suppfirst}

\begingroup
  \hypersetup{linkcolor=black}
  \tableofcontents
\endgroup
\clearpage

\section{Extended Related Work}
\label{app:related}

Existing compositional generation methods intervene at different stages of generation. They modify text representations, attention maps, denoising predictions, latent variables, model parameters, or the initial noise. We review these methods by the component they modify and the stage at which they intervene.

\subsection{Denoising-Time Attention Guidance and Semantic Binding}

Cross-attention links prompt tokens to spatial image features~\citep{vaswani2017attention,rombach2022ldm}. It has been used to analyze semantic correspondence~\citep{hertz2023prompt2prompt,tang2023daam} and to correct generation at inference time. Attend-and-Excite detects subject tokens with weak peak activations and repeatedly updates the current latent until every requested subject receives a strong response~\citep{chefer2023attend}. It mainly addresses concept omission. A-STAR separates subject attention maps while keeping each subject active during denoising~\citep{agarwal2023astar}. CONFORM uses contrastive objectives on token attention maps to improve subject presence and attribute correspondence~\citep{meral2024conform}. Other methods focus on semantic binding. SynGen first parses syntactic relation, then aligns the attention of related entities and modifiers and separates unrelated pairs~\citep{rassin2023sygen}. Divide~\&~Bind uses an attendance loss to strengthen object responses and a binding loss to align entities with their attributes~\citep{li2024dividebind}. Magnet instead constructs positive and negative binding vectors that modify the text embeddings~\citep{zhuang2024magnet}. ToMe merges related tokens into composite representations and refines them with objectives applied during early inference~\citep{hu2024tome}. These methods use different attention patterns to address concept omission and incorrect binding. Several approaches modify denoising predictions instead. Composable Diffusion combines concept-specific conditional scores using conjunction and negation operators~\citep{liu2022composable}. Structured Diffusion incorporates linguistic phrase structure into cross-attention~\citep{feng2023structure}. TweedieMix composes multiple concepts in Tweedie space~\citep{kwon2024tweediemix}. CO3 treats compositional failure as mode collision. It compares the joint prompt with single-concept prompts and suppresses mixed modes during early sampling~\citep{park2025co3}. All these methods intervene after sampling begins. RTD instead measures prompt-conditioned attention at the initial latent and applies one update before sampling. Standard denoising then proceeds without further correction.

\subsection{Layout- and Region-Based Compositional Control}

Layout-guided methods explicitly specify the desired spatial arrangement. Training-Free Layout Control guides cross-attention toward a user-provided layout~\citep{chen2024layout}. BoxDiff converts boxes or scribbles into spatial constraints and updates the latent during denoising to satisfy them~\citep{xie2023boxdiff}. DenseDiffusion adjusts cross-attention and self-attention according to segmentation layouts and compensates for differences in region size~\citep{kim2023densediffusion}. These methods provide coordinate-level control over object placement. Other methods generate the spatial plan automatically. LLM-Grounded Diffusion asks a large language model to produce captioned boxes and uses them to guide a pretrained diffusion model~\citep{lian2024llmgrounded}. MIGC introduces region-aware components for multi-instance generation~\citep{zhou2024migc}. \citet{wang2024attentioncontrol} predict entity regions with a learned BoxNet and use the resulting masks to control cross-attention and self-attention throughout sampling. UCC obtains a spatial and semantic plan from an LLM and uses it to guide early denoising~\citep{liu2026ucc}. Explicit or predicted layouts are useful when a prompt requires specific coordinates, counts, or relative positions. RTD does not use a target layout. It only encourages different concepts to receive distinct initial support. The pretrained generator still determines object scale, arrangement, and interaction.

\newpage
\subsection{Initial-Condition Optimization}

Diffusion outputs depend strongly on the sampled initial latent, which makes the latent a natural target for optimization before denoising. InitNO evaluates each candidate using a cross-attention response score, a self-attention conflict score, and a distribution alignment criterion~\citep{guo2024initno}. It iteratively updates the noise until all validity conditions are satisfied. RTD also modifies the initial latent, but it uses a single objective and one update. SOD measures competition between every pair of normalized concept maps using soft intersection over union. IGR follows the normalized separation gradient within a relative $\ell_2$ budget. InitNO therefore searches for an initial latent that satisfies several quality criteria. RTD instead applies one bounded local correction based on pairwise concept overlap and then runs the original sampler without further optimization.

\subsection{Multi-Concept Personalization and Attention Disentanglement}

Multi-concept personalization addresses a related problem in which several novel identities must be separated using reference images. DisenDiff learns concept-specific modifiers and calibrates attention to strengthen each concept while reducing interference~\citep{zhang2024disendiff}. AttenCraft extracts concept masks from self-attention and cross-attention and uses them to supervise disentanglement during customization~\citep{shentu2024attencraft}. ConceptSplit combines token-level value adaptation with latent optimization at inference time to preserve multiple learned identities~\citep{lim2025conceptsplit}. Both personalization and prompt composition must reduce attention entanglement, but they use different information. Personalization methods learn new identities from reference images and encode them in tokens or model parameters. RTD only uses the prompt and sampled latent to separate textual concepts already represented by the frozen generator.

\subsection{Structural Comparison}

\begin{table}[!h]
\centering
\caption{Structural comparison of representative compositional-generation methods. Auxiliary signals denote information beyond full-prompt conditioning. Training denotes model parameter updates. Layout denotes explicit spatial specification. Repeated denotes correction at multiple iterations.}
\label{tab:method_positioning}
\footnotesize
\begin{tabularx}{\linewidth}{>{\raggedright\arraybackslash}p{0.18\linewidth} >{\raggedright\arraybackslash}p{0.19\linewidth} >{\centering\arraybackslash}p{0.09\linewidth} >{\centering\arraybackslash}p{0.09\linewidth} >{\raggedright\arraybackslash}p{0.22\linewidth} C}
\toprule
\rowcolor{rtdgray}
\textbf{Method} & \textbf{Auxiliary signal} & \textbf{Training} & \textbf{Layout} & \textbf{Intervention point} & \textbf{Repeated} \\
\midrule
Attend-and-Excite~\citep{chefer2023attend} & Subject tokens & \xmark & \xmark & Denoising latent & \cmark \\
SynGen~\citep{rassin2023sygen} & Syntactic parse & \xmark & \xmark & Denoising latent & \cmark \\
BoxDiff~\citep{xie2023boxdiff} & Boxes / scribbles & \xmark & \cmark & Denoising latent & \cmark \\
LLM-Grounded~\citep{lian2024llmgrounded} & LLM-generated boxes & \xmark & \cmark & Planning + denoising & \cmark \\
InitNO~\citep{guo2024initno} & None & \xmark & \xmark & Initial noise & \cmark \\
CO3~\citep{park2025co3} & Single-concept prompts & \xmark & \xmark & Early denoising & \cmark \\
DisenDiff~\citep{zhang2024disendiff} & Reference image & \cmark & \xmark & Training + inference & \cmark \\
ConceptSplit~\citep{lim2025conceptsplit} & Reference images & \cmark & \xmark & Training + initial latent & \cmark \\
\rowcolor{rtdtealpale}
\textbf{RTD (ours)} & \textbf{Target concept phrases} & \xmark & \xmark & \textbf{Initial latent only} & \xmark \\
\bottomrule
\end{tabularx}
\end{table}

Table~\ref{tab:method_positioning} compares the supervision used by each method, its intervention point, and whether it applies repeated correction. We highlight that the proposed RTD is training-free and only modifies the initial latent. Its target concepts come from the prompt, and the pretrained network remains frozen.
The rectification objective is discarded before standard denoising begins. RTD therefore focuses on competition among textual concepts already represented by the generator. Such training-free methods are important because they eliminate task-specific training cost, which is critical in annotation-limited settings such as cold-start active learning~\citep{zhu2025csal,ma2025sugfw,zhu2025medcal} and label-scarce scenarios such as unsupervised anomaly detection~\citep{Ren2025deep,zhu2025frequency,Peng2025unsupervised,Zhu2024adversarial}.

\section{Method Details}
\label{app:method}
In this section, we provide additional implementation details. RTD takes a prompt $C$, a set of target concepts $\mathcal{C}=\{c_k\}_{k=1}^{K}$, and an initial latent $x_T$. The denoising network $\epsilon_\theta$ remains frozen during rectification and generation. This section describes \textsc{PilotAttention}, \textsc{Normalize}, and IGR in Algorithm~1 of the main paper.

\subsection{Concept Identification and Token Alignment}
\label{app:tokens}

Each target concept $c_k$ corresponds to one or more consecutive tokens in the full prompt. We denote their indices under the first SDXL tokenizer by $\mathcal{I}_k\subseteq\{1,\ldots,L\}$. We tokenize both $C$ and $c_k$, remove padding and special tokens, and normalize tokenizer-specific continuation markers. We then locate the first exact occurrence of the concept token sequence in the full prompt. The matched indices select the cross-attention columns for $c_k$. We obtain concept phrases according to each benchmark protocol. For AE-Bench, we parse the entity slots in the templated prompts using whitespace. For T2I-CompBench, we use the provided entity annotations. For RareBench, we extract noun phrases with Stanza~\citep{qi2020stanza}. Concept extraction only identifies the textual entities to compare. It provides no object locations or masks. When a phrase contains multiple tokens, we sum their attention maps to obtain one concept map.

\subsection{Pilot Cross-Attention Aggregation}
\label{app:pilot_aggregation}

The pilot uses the first timestep of the 50-step DDIM schedule, $t_{\mathrm{pilot}}=980$. RTD performs one prompt-conditioned U-Net pass at $x_T$ and records cross-attention probabilities from modules with a $32\times32$ spatial grid. Within each module, we average the probabilities across attention heads and sum the columns indexed by $\mathcal{I}_k$. We then average the resulting concept maps across the recorded modules to obtain $A^{(k)}$. Using one spatial resolution avoids mixing maps with different granularities. Averaging across heads and modules reduces isolated responses and retains spatial evidence shared across the network. The pilot pass is diagnostic only. It does not call the scheduler or advance $x_T$ to a lower-noise state.

\subsection{Soft-Overlap Disentanglement}
\label{app:sod_details}

Raw attention magnitudes are not directly comparable across concepts. SOD applies min-max normalization to each $A^{(k)}$ to obtain a soft occupancy map $M^{(k)}\in[0,1]^{32\times32}$. It then computes soft intersection over union for every concept pair and maximizes their mean separation:
\begin{equation}
\begin{aligned}
M^{(k)}_p
&=\frac{A^{(k)}_p-\min_q A^{(k)}_q}
        {\max_q A^{(k)}_q-\min_q A^{(k)}_q+\varepsilon},\\
O_{ij}
&=\frac{\langle M^{(i)},M^{(j)}\rangle}
        {\|M^{(i)}\|_1+\|M^{(j)}\|_1
        -\langle M^{(i)},M^{(j)}\rangle+\varepsilon},\\
\mathcal{S}(x_T)
&=1-\frac{2}{K(K-1)}\sum_{1\leq i<j\leq K}O_{ij},
\qquad \varepsilon=10^{-6}.
\end{aligned}
\label{eq:app_sod_compact}
\end{equation}
Here, $p$ and $q$ index spatial cells, and $K\geq2$ for the compositional prompts considered in this work. Normalization preserves the spatial ordering within each map while removing differences in response scale. The resulting objective is differentiable almost everywhere with respect to the attention values. The factor $2/[K(K-1)]$ gives equal weight to each concept pair and keeps the objective comparable across different numbers of concepts.

\subsection{Isotropic Gradient Rectification}
\label{app:igr_details}

IGR differentiates Eq.~\eqref{eq:app_sod_compact} with respect to $x_T$ while keeping all network parameters frozen. Let $\delta=x_T'-x_T$ denote a correction to the initial latent and let $B=\rho\|x_T\|_2$ be its prescribed budget. A first-order expansion gives
\begin{equation}
    \mathcal{S}(x_T+\delta)
    \approx
    \mathcal{S}(x_T)+g^\top\delta,
    \qquad
    g=\nabla_{x_T}\mathcal{S}(x_T).
\label{eq:app_igr_linearization}
\end{equation}

The first-order optimal correction within the $\ell_2$ neighborhood solves
\begin{equation}
    \delta^\star
    =
    \underset{\|\delta\|_2\leq B}{\argmax}\,
    g^\top\delta.
\label{eq:app_igr_constrained}
\end{equation}
By the Cauchy Schwarz inequality,
$g^\top\delta\leq\|g\|_2\|\delta\|_2\leq B\|g\|_2$.
For a nonzero gradient, equality is attained when $\delta$ is aligned with $g$ and uses the full budget. Consequently,
\begin{equation}
    \delta^\star
    =
    \rho\|x_T\|_2\frac{g}{\|g\|_2},
    \qquad
    x_T'=x_T+\delta^\star.
\label{eq:app_igr_solution}
\end{equation}

For numerical stability, RTD implements this solution as
\begin{equation}
    \widehat g=\frac{g}{\max(\|g\|_2,10^{-8})},
    \qquad
    x_T'=x_T+\rho\|x_T\|_2\widehat g,
    \quad \rho=0.02.
\label{eq:app_igr_compact}
\end{equation}
The $\ell_2$ constraint gives every latent direction the same correction budget, which motivates the term \emph{isotropic}. Gradient normalization makes the update independent of the raw gradient magnitude, while $\rho$ sets the displacement relative to the norm of the sampled latent. When $\|g\|_2\geq10^{-8}$, Eq.~\eqref{eq:app_igr_compact} matches the exact solution to the linearized constrained problem. For a smaller gradient, the numerical safeguard reduces the update continuously toward zero. After the update, RTD discards the SOD graph and passes $(x_T',C)$ to the original sampler without further intervention.

\section{Experimental and Reproducibility Details}
\label{app:experiments}

\subsection{Generation and Computing Configuration}
\label{app:generation_config}

The main experiments use \href{https://huggingface.co/stabilityai/stable-diffusion-xl-base-1.0}{Stable Diffusion XL Base 1.0} with the \href{https://huggingface.co/madebyollin/sdxl-vae-fp16-fix}{SDXL VAE fix}. We load both tokenizer and text encoder pairs from the SDXL checkpoint. Text inputs use maximum-length padding and the default truncation rule. We use the native DDIM scheduler with deterministic sampling. Images are generated at $1024\times1024$ resolution with 50 inference steps, classifier-free guidance at scale $5.0$, and an empty negative prompt. RTD runs the pilot at $t_{\mathrm{pilot}}=980$, which is the first timestep in this schedule, and records cross-attention from all modules with a $32\times32$ spatial resolution. We average attention across heads and modules and sum the subword token maps for each concept as described in Section~\ref{app:pilot_aggregation}. IGR applies one update with $\rho=0.02$. DDIM sampling then starts from the rectified latent without further RTD intervention. The environment uses Python 3.10.20, PyTorch 2.5.1, and CUDA 12.1. All main experiments run on one NVIDIA H200 GPU. RTD stores no additional parameters and requires no distributed generation or model parallelism. Vanilla generation and RTD therefore use the same SDXL checkpoint. This configuration applies to RTD and the controlled SDXL comparisons. For each baseline, we use the guidance rule and method-specific hyperparameters from its official implementation. Prompts and other evaluation procedures
 remain fixed in paired comparisons.

\subsection{Benchmark Protocol}

For AE-Bench, we report BLIP-VQA and ImageReward separately for its three categories. For T2I-CompBench and RareBench, we report ImageReward and the human evaluation score described below. Each generated image is evaluated against its corresponding benchmark prompt, and all metrics for a method are computed on the same generated set.

\begin{table}[H]
\centering
\caption{Benchmark subsets and concept sources used in the main paper.}
\label{tab:benchmark_protocol}
\small
\begin{tabularx}{\linewidth}{>{\raggedright\arraybackslash}p{0.18\linewidth} >{\raggedright\arraybackslash}p{0.27\linewidth} >{\centering\arraybackslash}p{0.10\linewidth} Y}
\toprule
\rowcolor{rtdgray}
\textbf{Benchmark} & \textbf{Evaluated split} & \textbf{Prompts} & \textbf{Target-concept source} \\
\midrule
AE-Bench~\citep{chefer2023attend} & Animal-animal, animal-object, object-object & 276 & Whitespace-parsed entity slots in the benchmark's templated prompts \\
T2I-CompBench~\citep{huang2023t2icompbench} & Complex compositional split & 1,000 & Entity annotations distributed with the benchmark \\
RareBench~\citep{park2025co3,kim2025r2f} & Concat, Relations, Complex & 120 & English noun phrases extracted with Stanza \\
\bottomrule
\end{tabularx}
\end{table}

\paragraph{Human evaluation protocol}
For each prompt, we present the outputs from RTD and the baselines side by side. Method names are hidden, and the display order is randomized independently. All 11 participants evaluate the output from every seed for each prompt and method. They rate prompt to image agreement on an integer scale from 0 to 5, where a larger value indicates better agreement. We divide each rating $s$ by 5 and average the normalized ratings across participants and seeds to obtain the final score in $[0,1]$.

\subsection{Dataset and Model Sources}
\label{app:sources}

Tables~\ref{tab:data_sources} and~\ref{tab:software_sources} list the public resources used for prompts, checkpoints, evaluation, and baseline implementations.

\begin{table}[H]
\centering
\caption{Public benchmark and evaluation sources.}
\label{tab:data_sources}
\small
\begin{tabularx}{\linewidth}{>{\raggedright\arraybackslash}p{0.15\linewidth} >{\raggedright\arraybackslash}p{0.49\linewidth} Y}
\toprule
\rowcolor{rtdgray}
\textbf{Resource} & \textbf{Official website} & \textbf{Use in this paper} \\
\midrule
AE-Bench & \url{https://github.com/yuval-alaluf/Attend-and-Excite} & Prompt set and BLIP-VQA protocol \\
T2I-CompBench & \url{https://github.com/Karine-Huang/T2I-CompBench} & Complex prompts and annotations \\
RareBench & \url{https://github.com/krafton-ai/Rare-to-Frequent} & Long-tail prompt subsets \\
ImageReward & \url{https://github.com/zai-org/ImageReward} & Text-to-image preference score \\
BLIP-2 / VQA & \url{https://github.com/salesforce/LAVIS} & AE-Bench concept-presence score \\
Stanza & \url{https://github.com/stanfordnlp/stanza} & RareBench noun-phrase parsing \\
\bottomrule
\end{tabularx}
\end{table}

\begin{table}[H]
\centering
\caption{Official model and representative baseline implementation sources.}
\label{tab:software_sources}
\footnotesize
\begin{tabularx}{\linewidth}{>{\raggedright\arraybackslash}p{0.23\linewidth} Y}
\toprule
\rowcolor{rtdgray}
\textbf{Component} & \textbf{Official website} \\
\midrule
SDXL & \url{https://huggingface.co/stabilityai/stable-diffusion-xl-base-1.0} \\
SDXL VAE fix & \url{https://huggingface.co/madebyollin/sdxl-vae-fp16-fix} \\
Diffusers & \url{https://github.com/huggingface/diffusers} \\
Attend-and-Excite & \url{https://github.com/yuval-alaluf/Attend-and-Excite} \\
SynGen & \url{https://github.com/RoyiRa/Linguistic-Binding-in-Diffusion-Models} \\
Divide~\&~Bind & \url{https://sites.google.com/view/divide-and-bind/startseite} \\
Composable Diffusion & \url{https://github.com/energy-based-model/Compositional-Visual-Generation-with-Composable-Diffusion-Models-PyTorch} \\
TweedieMix & \url{https://github.com/KwonGihyun/TweedieMix} \\
Magnet & \url{https://github.com/I2-Multimedia-Lab/Magnet} \\
ToMe & \url{https://github.com/hutaihang/ToMe} \\
R2F & \url{https://github.com/krafton-ai/Rare-to-Frequent} \\
InitNO & \url{https://github.com/xiefan-guo/initno} \\
CO3 & \url{https://github.com/debottam-dutta7/co3} \\
\bottomrule
\end{tabularx}
\end{table}

\section{Additional Qualitative Results and Analysis}
\label{app:additional_qualitative}

We compare CO3, R2F, ToMe, and RTD under the setting used in the main paper. The six panels cover semantically similar entities, object pairs with attributes, spatial relations, unusual materials and shapes, and long-tail interactions. These examples complement the automatic and human evaluations by showing how each method represents the concepts in a prompt.

\begin{figure}[p]
    \centering
    \includegraphics[width=\textwidth]{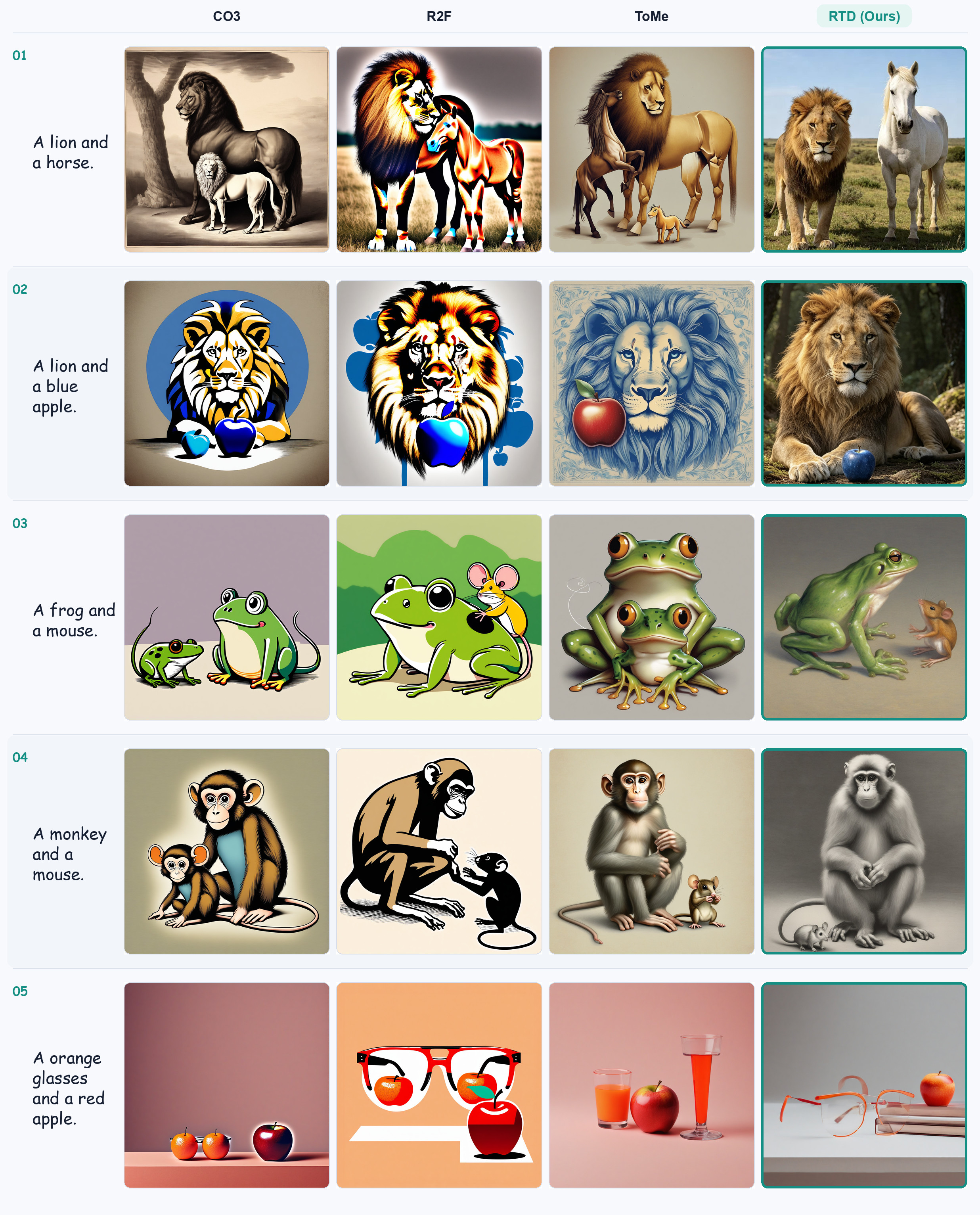}
    \caption{\textbf{AE-Bench, examples 1 to 5.} Rows correspond to prompts. Columns show CO3, R2F, ToMe, and RTD from left to right. The examples emphasize category separation for semantically related subjects and unequal object scales.}
    \label{fig:appendix_qualitative_ae_top}
\end{figure}
\begin{figure}[p]
    \centering
    \includegraphics[width=\textwidth]{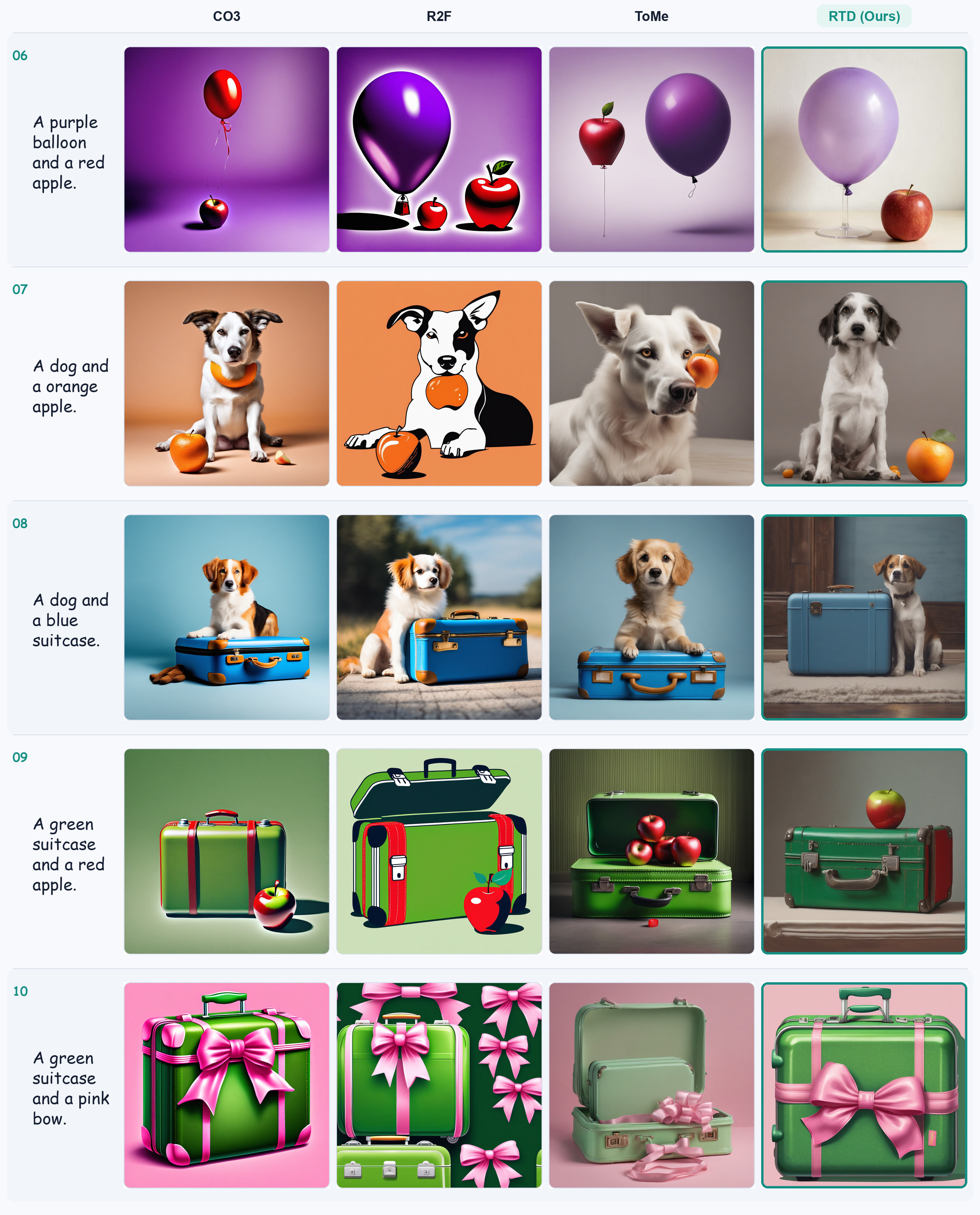}
    \caption{\textbf{AE-Bench, examples 6 to 10.} Rows correspond to prompts. Columns show CO3, R2F, ToMe, and RTD from left to right. These cases combine object identity with color attributes.}
    \label{fig:appendix_qualitative_ae_bottom}
\end{figure}

\begin{figure}[p]
    \centering
    \includegraphics[width=\textwidth]{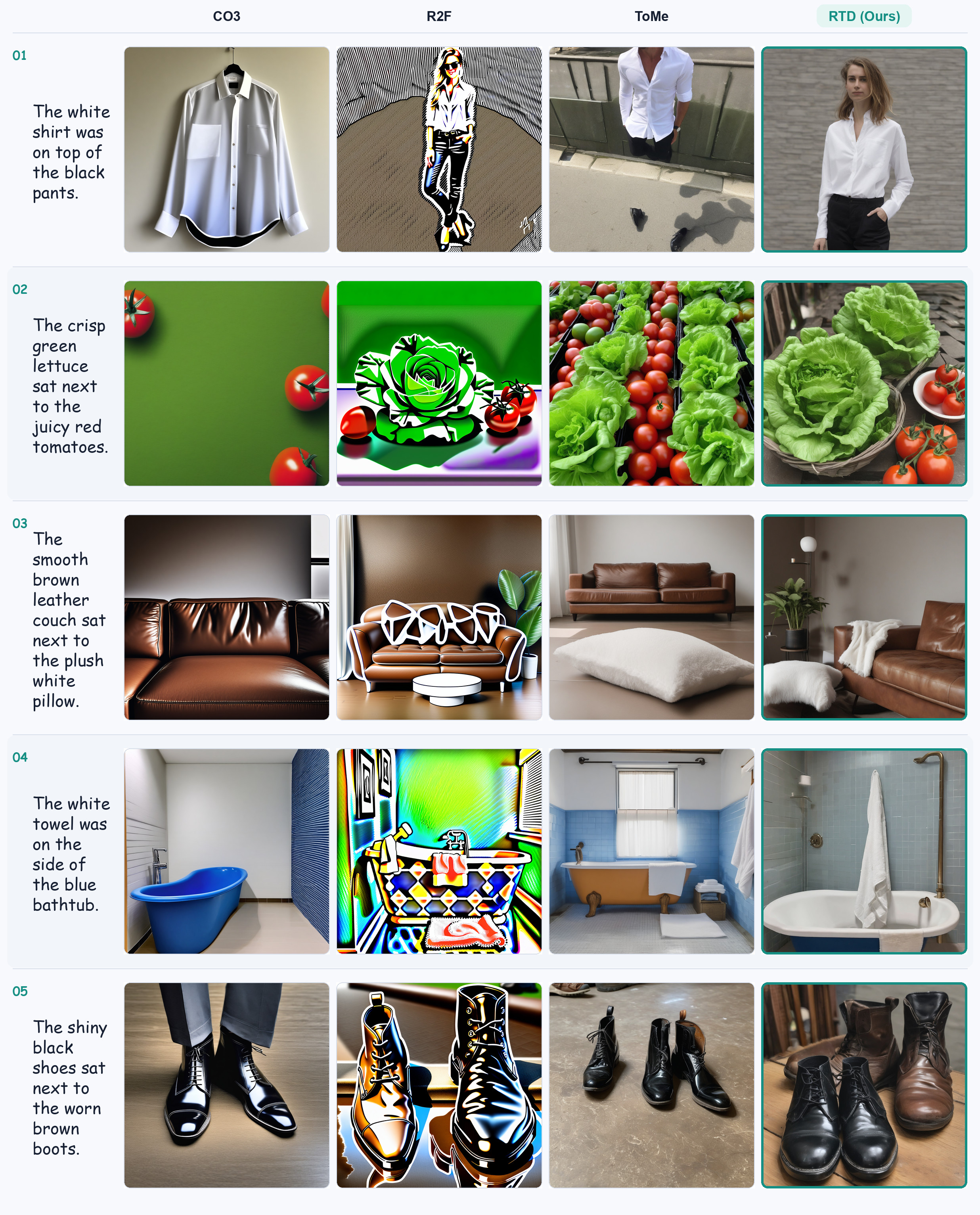}
    \caption{\textbf{T2I-CompBench, examples 1 to 5.} Rows correspond to prompts. Columns show CO3, R2F, ToMe, and RTD from left to right. The cases combine entity identity with attributes and local spatial relations, including adjacent or contacting objects.}
    \label{fig:appendix_qualitative_t2i_top}
\end{figure}
\begin{figure}[p]
    \centering
    \includegraphics[width=\textwidth]{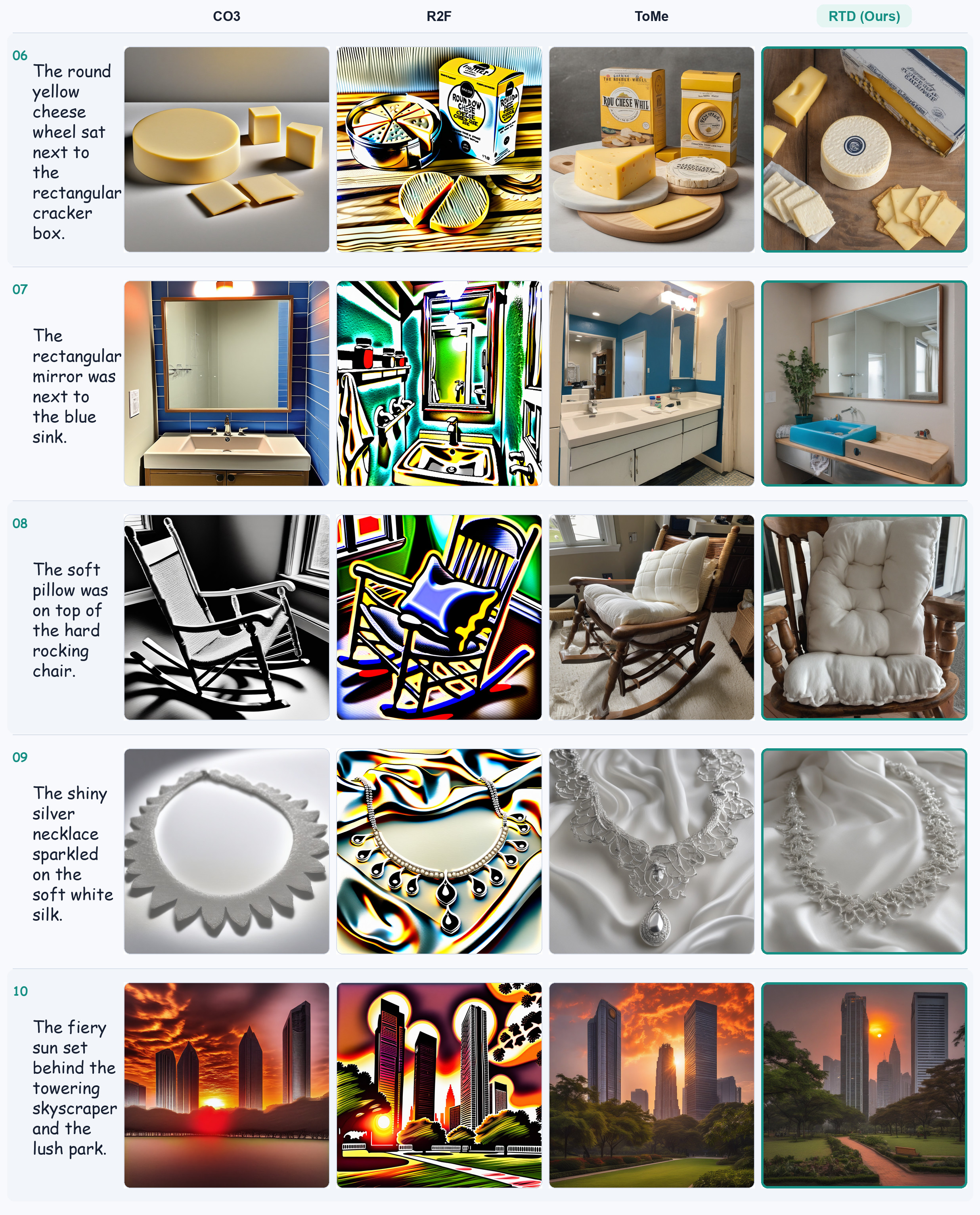}
    \caption{\textbf{T2I-CompBench, examples 6 to 10.} Rows correspond to prompts. Columns show CO3, R2F, ToMe, and RTD from left to right. The cases include contact, material, multi-object, and scene-level compositions. Color, package identity, and object count remain challenging in several examples.}
    \label{fig:appendix_qualitative_t2i_bottom}
\end{figure}

\begin{figure}[p]
    \centering
    \includegraphics[width=\textwidth]{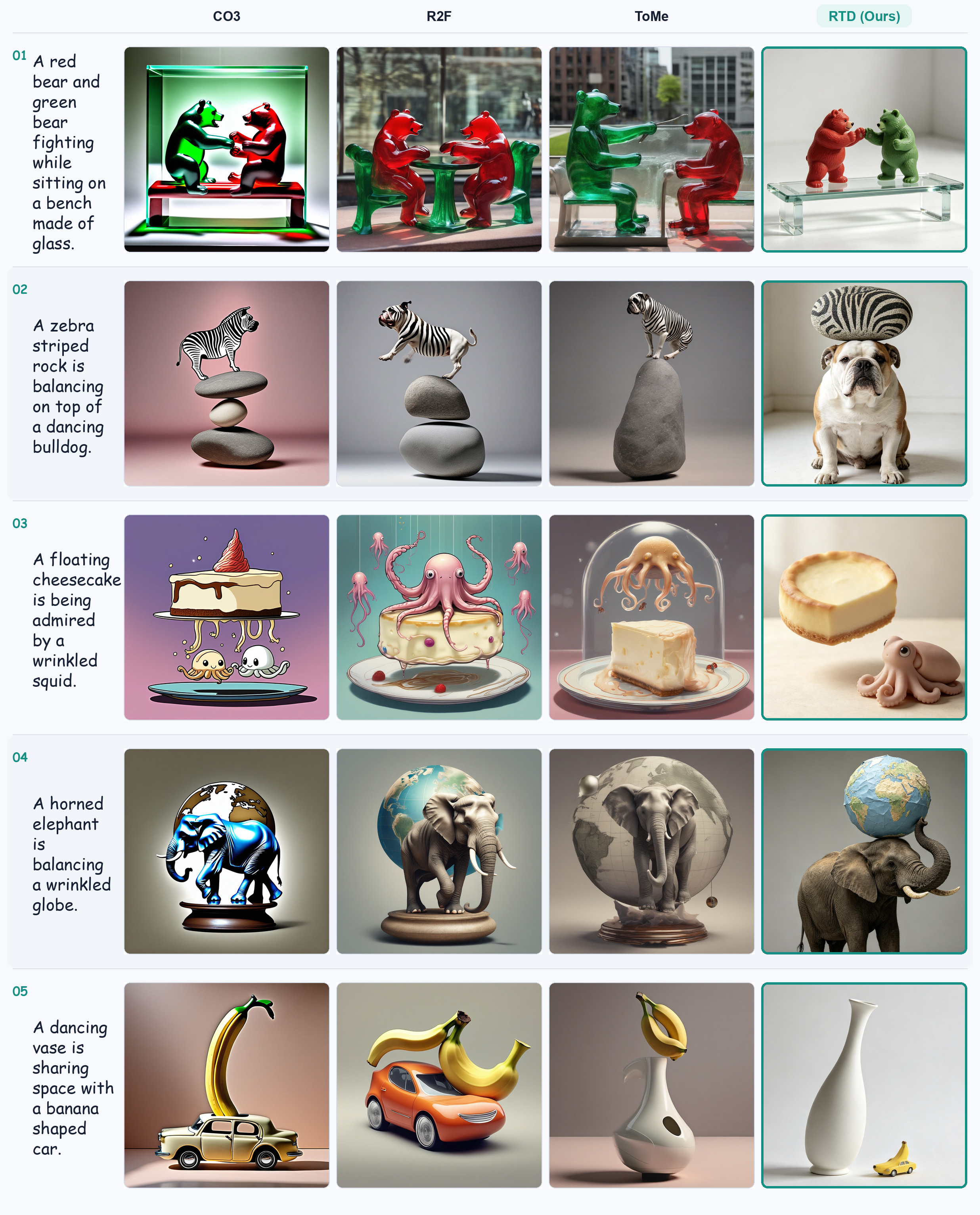}
    \caption{\textbf{RareBench, examples 1 to 5.} Rows correspond to prompts. Columns show CO3, R2F, ToMe, and RTD from left to right. The cases emphasize uncommon attributes, materials, and multi-concept relations.}
    \label{fig:appendix_qualitative_rare_top}
\end{figure}
\begin{figure}[p]
    \centering
    \includegraphics[width=\textwidth]{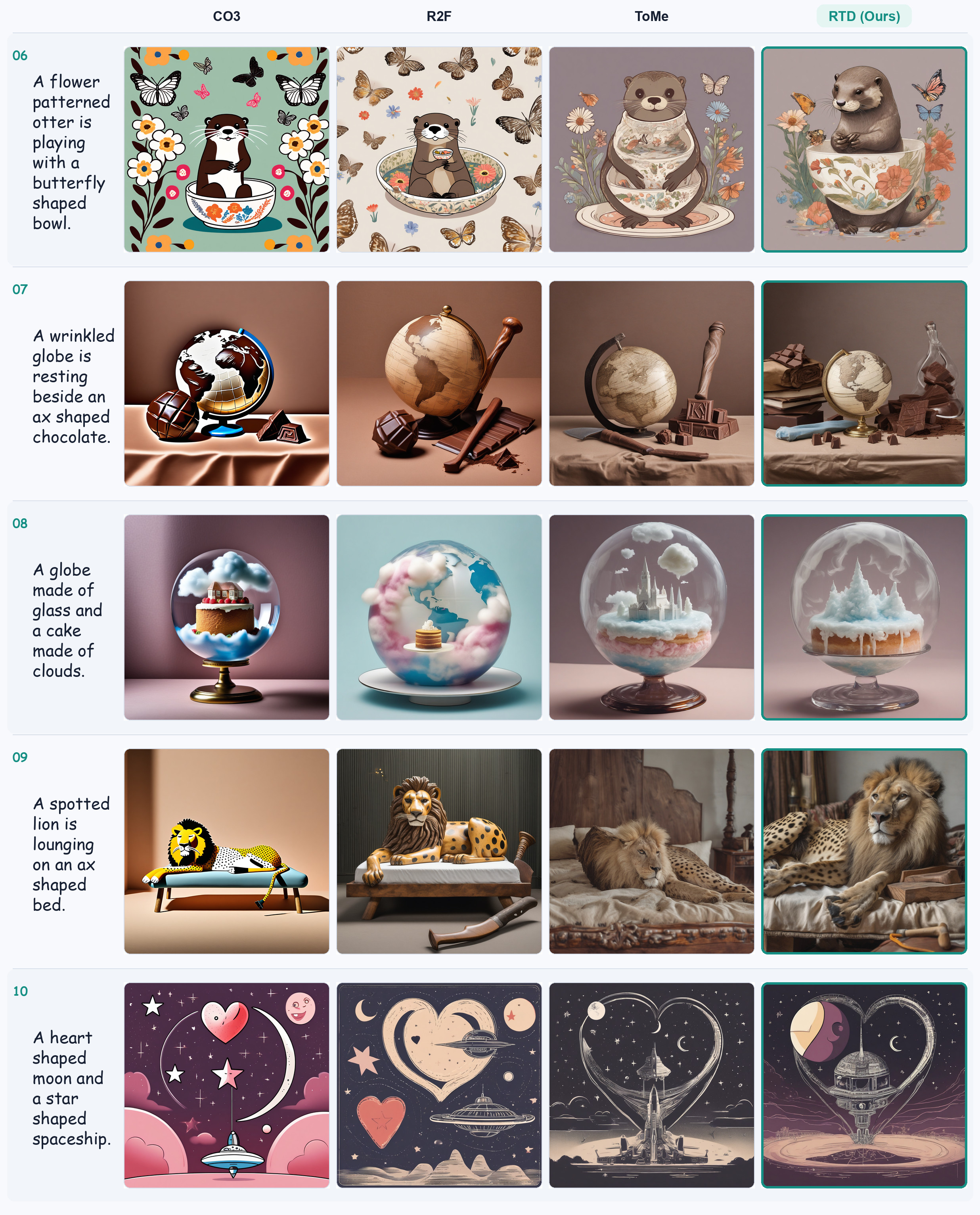}
    \caption{\textbf{RareBench, examples 6 to 10.} Rows correspond to prompts. Columns show CO3, R2F, ToMe, and RTD from left to right. These cases focus on phrase-internal shape, material composition, and interactions.}
    \label{fig:appendix_qualitative_rare_bottom}
\end{figure}
\FloatBarrier

\subsection{Observed Scope and Failure Modes}
\label{app:qualitative_analysis}

We first separate successful generations from failure cases. Table~\ref{tab:qualitative_success} summarizes examples in which RTD represents all target concepts and preserves the requested attributes or relations. Table~\ref{tab:qualitative_failure} reports examples in which the target entities are partly present but a finer prompt constraint is not satisfied.

\begin{center}
\captionsetup{hypcap=false}
\captionof{table}{Representative successful generations by RTD.}
\label{tab:qualitative_success}
\footnotesize
\begin{tabularx}{\linewidth}{L{0.20\linewidth} L{0.38\linewidth} Y}
\toprule
\rowcolor{rtdgray}
\textbf{Capability} & \textbf{Representative cases} & \textbf{Visual outcome} \\
\midrule
Concept coverage & Lion and horse (Fig.~\ref{fig:appendix_qualitative_ae_top}, Ex.~1), frog and mouse (Fig.~\ref{fig:appendix_qualitative_ae_top}, Ex.~3), shirt and pants (Fig.~\ref{fig:appendix_qualitative_t2i_top}, Ex.~1) & Both requested concepts appear as identifiable entities. \\
Attribute binding & Blue apple (Fig.~\ref{fig:appendix_qualitative_ae_top}, Ex.~2), blue suitcase (Fig.~\ref{fig:appendix_qualitative_ae_bottom}, Ex.~8), zebra-striped rock (Fig.~\ref{fig:appendix_qualitative_rare_top}, Ex.~2) & The attributes are attached to the intended objects. \\
Spatial relation & Couch and pillow (Fig.~\ref{fig:appendix_qualitative_t2i_top}, Ex.~3), mirror and sink (Fig.~\ref{fig:appendix_qualitative_t2i_bottom}, Ex.~7), elephant and globe (Fig.~\ref{fig:appendix_qualitative_rare_top}, Ex.~4) & The requested adjacency or vertical relation is preserved. \\
Shape and interaction & Banana-shaped car (Fig.~\ref{fig:appendix_qualitative_rare_top}, Ex.~5), fighting bears (Fig.~\ref{fig:appendix_qualitative_rare_top}, Ex.~1), cheesecake admired by a squid (Fig.~\ref{fig:appendix_qualitative_rare_top}, Ex.~3) & The requested shape or interaction is expressed while the participating concepts remain distinct. \\
\bottomrule
\end{tabularx}
\end{center}

\begin{center}
\captionsetup{hypcap=false}
\captionof{table}{Representative failure cases of RTD.}
\label{tab:qualitative_failure}
\footnotesize
\begin{tabularx}{\linewidth}{L{0.20\linewidth} L{0.38\linewidth} Y}
\toprule
\rowcolor{rtdgray}
\textbf{Failure type} & \textbf{Representative case} & \textbf{Observed failure} \\
\midrule
Action execution & Zebra-striped rock on a dancing bulldog (Fig.~\ref{fig:appendix_qualitative_rare_top}, Ex.~2) & The rock and bulldog appear, but the bulldog does not express a dancing pose. \\
Modifier realization & Ax-shaped chocolate (Fig.~\ref{fig:appendix_qualitative_rare_bottom}, Ex.~7), ax-shaped bed (Fig.~\ref{fig:appendix_qualitative_rare_bottom}, Ex.~9) & The ax may appear as a separate element instead of determining the shape of the chocolate or bed. \\
Fine-grained identity & Cheese wheel next to a rectangular cracker box (Fig.~\ref{fig:appendix_qualitative_t2i_bottom}, Ex.~6) & The cheese wheel is clear, but the rectangular cracker box is not rendered unambiguously. \\
\bottomrule
\end{tabularx}
\end{center}

\paragraph{Why RTD succeeds}
SOD reduces overlap between the attention maps of complete target concepts. This objective directly addresses omission and fusion, which explains the consistent concept coverage in Table~\ref{tab:qualitative_success}. Once the concepts receive distinct initial support, the frozen generator can use its pretrained knowledge to produce their attributes, shapes, and relations. The banana-shaped car demonstrates this behavior. RTD separates the relevant concepts, while the generator supplies the visual form requested by the prompt.

\paragraph{Why RTD fails}
The failures in Table~\ref{tab:qualitative_failure} occur when concept separation alone is insufficient. SOD assigns one map to each complete target phrase and does not model the internal roles of nouns, modifiers, and actions. It can preserve a bulldog and a rock without enforcing a dancing pose. It can also preserve an ax and a bed without requiring the bed itself to take the shape of an ax. In addition, the pilot maps have a spatial resolution of $32\times32$ and are averaged across heads and modules. They capture broad concept allocation more directly than fine geometry, object identity, or contact. IGR applies one update before denoising and does not correct these details later in the trajectory.

\paragraph{Strengths and limitations}
RTD is most effective when failure is caused by competition between concepts. It improves concept coverage, reduces semantic fusion, and can preserve attributes and broad spatial relations without training, explicit layouts, or repeated sampling guidance. Its main limitation is that it optimizes concept allocation rather than complete scene semantics. Fine-grained identity, phrase-internal modifiers, exact pose, count, and contact remain dependent on the pretrained generator. RTD therefore provides a strong initial condition for compositional generation, but it does not replace explicit control of detailed attributes and relations.

\clearpage
\section{Additional Attention-Evolution Visualizations}
\label{app:attention_evolution}

\begin{figure}[!h]
    \centering
    \includegraphics[width=0.91\textwidth,height=0.85\textheight,keepaspectratio]{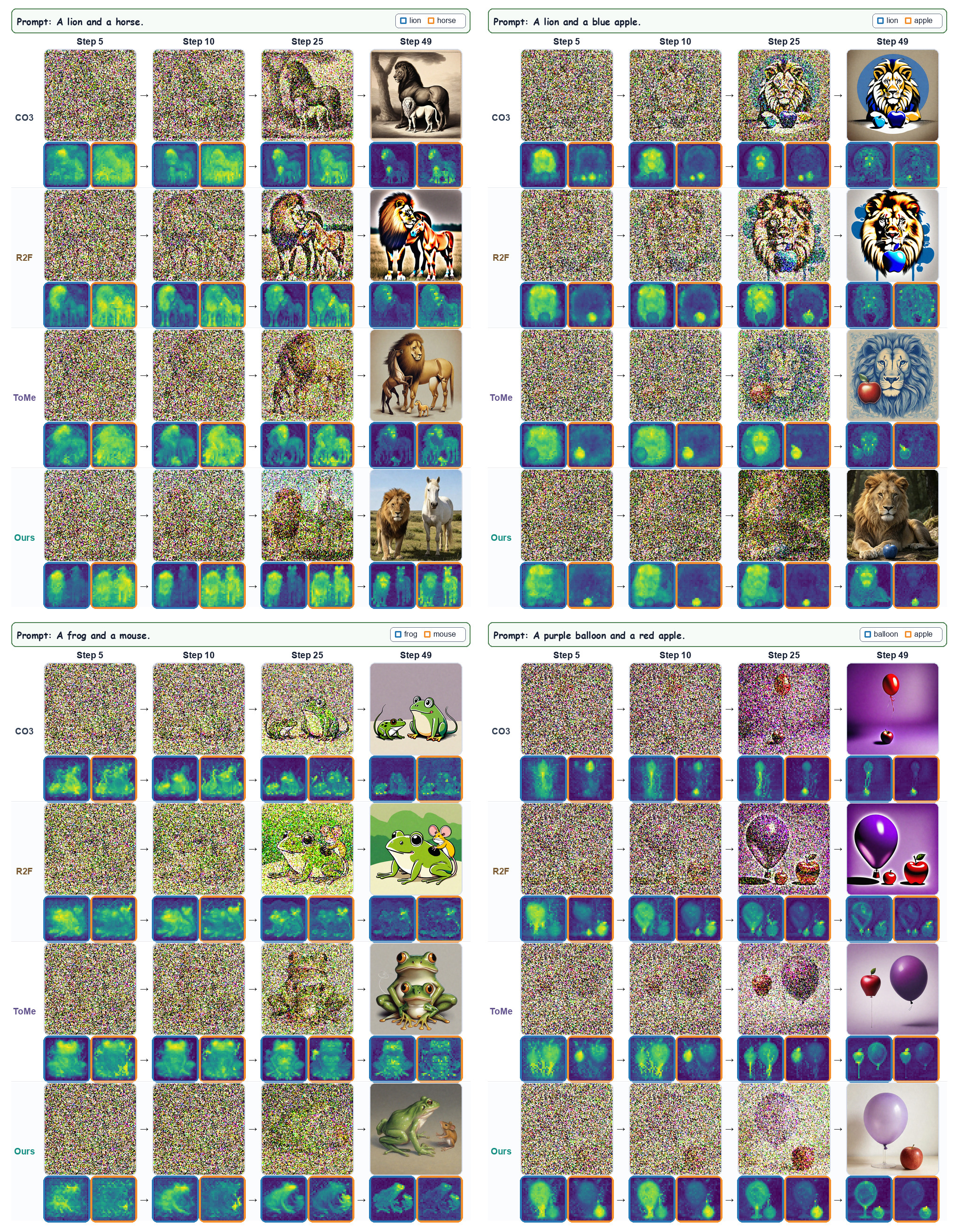}
    \caption{Denoising and attention evolution on AE-Bench. Within each prompt, rows show CO3, R2F, ToMe, and RTD. Columns show Steps 5, 10, 25, and 49. Each decoded state is followed by its concept maps. Border colors correspond to the legend.}
    \label{fig:appendix_attention_ae}
\end{figure}

\begin{figure}[p]
    \centering
    \includegraphics[width=\textwidth,height=\textheight,keepaspectratio]{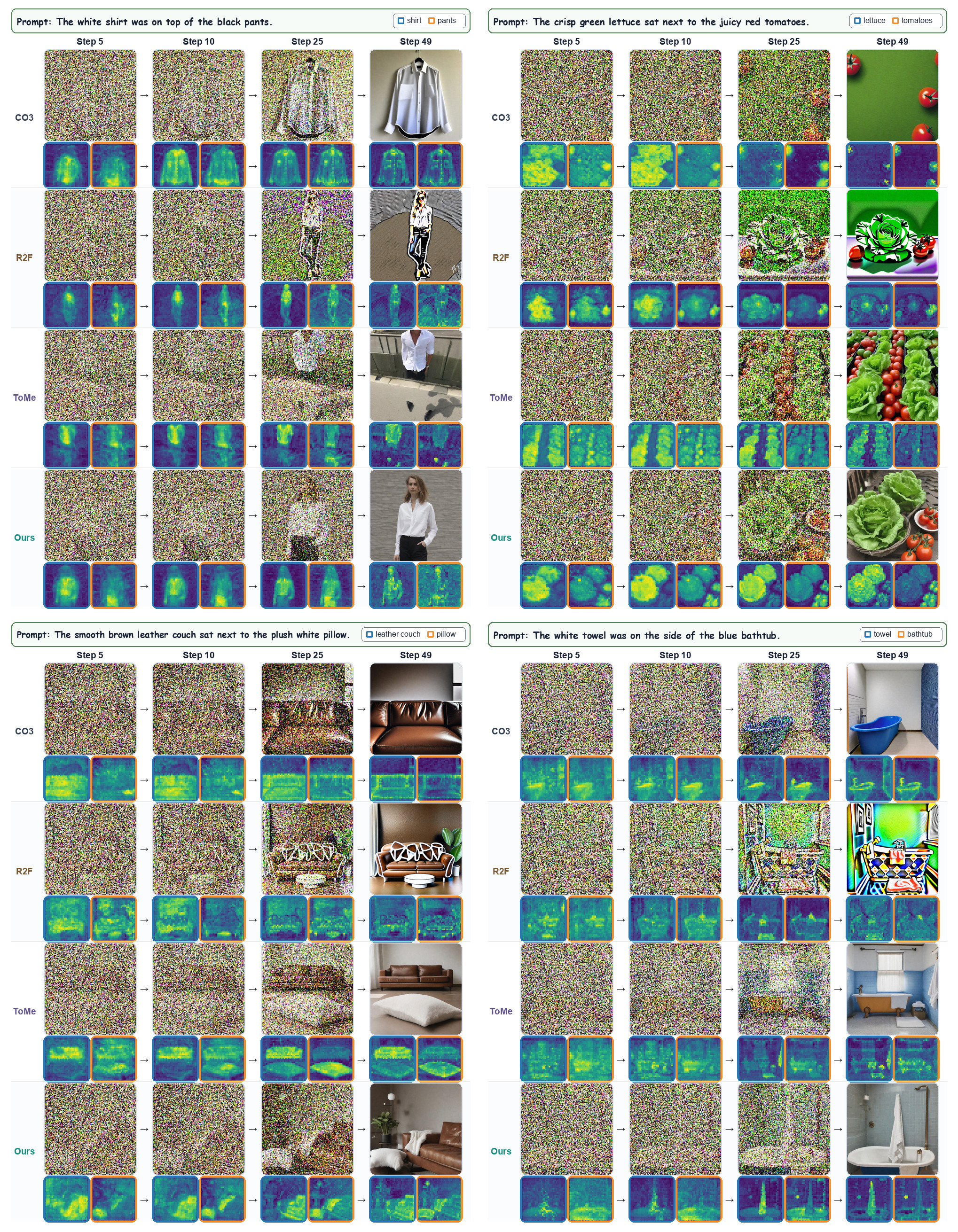}
    \caption{Denoising and attention evolution on T2I-CompBench. Within each prompt, rows show CO3, R2F, ToMe, and RTD. Columns show Steps 5, 10, 25, and 49. Each decoded state is followed by its concept maps. Border colors correspond to the legend.}
    \label{fig:appendix_attention_t2i}
\end{figure}

\begin{figure}[p]
    \centering
    \includegraphics[width=\textwidth,height=\textheight,keepaspectratio]{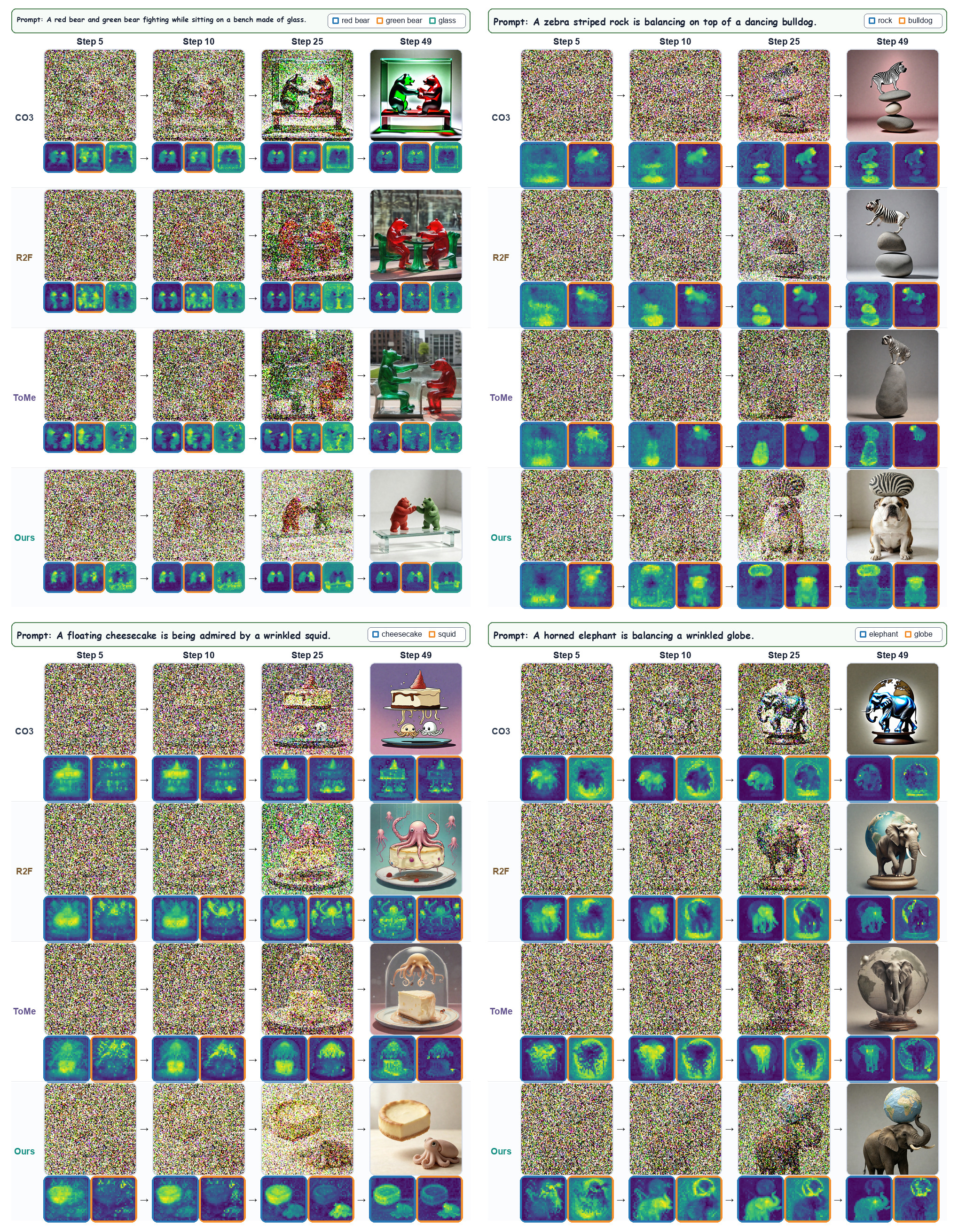}
    \caption{Denoising and attention evolution on RareBench. Within each prompt, rows show CO3, R2F, ToMe, and RTD. Columns show Steps 5, 10, 25, and 49. Each decoded state is followed by its concept maps. Border colors correspond to the legend.}
    \label{fig:appendix_attention_rare}
\end{figure}

\FloatBarrier
Figures~\ref{fig:appendix_attention_ae} to~\ref{fig:appendix_attention_rare} show decoded intermediate states and concept-specific cross-attention maps at Steps 5, 10, 25, and 49. Blue and orange frames mark the two target concepts. Teal marks the glass attribute in the red bear example. At Steps 5 and 10, the decoded states are still dominated by noise, but the maps already show prompt-dependent spatial support. Steps 25 and 49 show how this support develops into recognizable structure. RTD produces compact and distinct support early in the displayed trajectories. Prompts containing a lion and a horse or a frog and a mouse show separation between concepts with similar semantics or different scales. Prompts containing a shirt and pants, a couch and pillow, or a towel and bathtub show that the maps can remain distinct when the objects are adjacent or in contact. RareBench includes unusual modifiers and vertical relations, such as a zebra-striped rock, a floating cheesecake, and an elephant with a globe. In each case, SOD provides a direction that reduces pilot-map overlap, IGR applies one bounded update to the initial latent, and the unchanged sampler develops the modified support into visual structure. The benchmark results in the main paper report aggregate compositional fidelity.

\section{Limitations}
\label{app:limitations_impact}

\paragraph{Objective scope}
SOD treats each extracted concept phrase as one semantic unit. It does not model the separate roles of head nouns, modifiers, or relations. The pretrained model therefore still controls phrase-level shape, pose, object count, relative scale, occlusion, and exact contact. Separation alone may also be insufficient for relations that require spatial overlap, such as wearing or holding.

\paragraph{Dependence on the pretrained model}
RTD only reorganizes spatial support using knowledge in the frozen generator. It does not add visual knowledge or modify model parameters. Concepts that are poorly represented by the backbone may remain difficult after rectification. Results may also vary across prompts and initial samples because RTD applies only one local update.

\paragraph{Architectural and empirical scope}
The current implementation requires differentiable attention that aligns text tokens with spatial features. The experiments cover latent diffusion backbones, multiple solvers, and several sampling budgets. Architectures without a similar text-to-space interface remain untested. Multilingual prompts, larger concept sets, and specialized visual domains also require further study.

\FloatBarrier
\clearpage

\FloatBarrier
\twocolumn
\phantomsection
\addcontentsline{toc}{section}{References}
\bibliography{reference}

\end{document}